%% file: main.tex
\documentclass[sigconf,balance=false,table]{acmart}
\usepackage{popets}

\usepackage{multirow}
\usepackage{comment}
\usepackage{amsfonts}
\usepackage{graphicx}
\usepackage{ifthen}

\usepackage{shellesc}
\usepackage{xcolor}

\long\def\ignore#1{}
\definecolor{lightgreen}{HTML}{e5f4d5}
\definecolor{lightblue}{HTML}{a2d2ff}

\usepackage{popets}
\setcopyright{popets}
\copyrightyear{YYYY}
\acmYear{YYYY}
\acmVolume{YYYY}
\acmNumber{X}
\acmDOI{XXXXXXX.XXXXXXX}
\acmISBN{}
\acmConference{Proceedings on Privacy Enhancing Technologies}
\RequirePackage[normalem]{ulem} %DIF PREAMBLE
\RequirePackage{color}\definecolor{RED}{rgb}{1,0,0}\definecolor{BLUE}{rgb}{0,0,1} %DIF PREAMBLE
\begin{document}


\title{DiDOTS: Knowledge Distillation from Large-Language-Models for Dementia Obfuscation in Transcribed Speech}

%\title{AD-GUARD: Knowledge Distillation from Large-Language-Models to Obfuscate Dementia in Transcribed Speech}

%alternative:Obfuscating Dementia in Speech with Distilled LLMs
%alternative: Obfuscating Dementia in Speech with LLM Distillation

\author{Dominika Woszczyk}
\orcid{0000-0001-9589-1843}
\affiliation{%
  \institution{Imperial College London}
  \city{London}
  \state{}
  \country{UK}
}
\email{d.woszczyk19@imperial.ac.uk}

\author{Soteris Demetriou}
\orcid{0000-0003-0318-9171}
\affiliation{%
  \institution{Imperial College London}
  \city{London}
  \state{}
  \country{UK}
}
\email{s.demetriou@imperial.ac.uk}

\renewcommand{\shortauthors}{Woszczyk et al.}
\newcommand{\systemName}[1]{DiDOTS}

\input{Sections/abstract}


\maketitle

\ShellEscape{latexdiff Sections/introduction_submitted.tex Sections/introduction.tex > Sections/introduction-d.tex}

\ShellEscape{latexdiff Sections/motivation_submitted.tex Sections/motivation.tex > Sections/motivation-d.tex}

\ShellEscape{latexdiff Sections/method_submitted.tex Sections/method.tex > Sections/method-d.tex}

\ShellEscape{latexdiff Sections/discussion_submitted.tex Sections/discussion.tex > Sections/discussion-d.tex}

\ShellEscape{latexdiff Sections/evalSetup_submitted.tex Sections/evalSetup.tex > Sections/evalSetup-d.tex}

\ShellEscape{latexdiff Sections/evalResults_submitted.tex Sections/evalResults.tex > Sections/evalResults-d.tex}

\newboolean{render_diff}
\setboolean{render_diff}{false}

\ifthenelse{\boolean{render_diff}}
{
\input{Sections/introduction-d}
\input{Sections/motivation-d}
\input{Sections/method-d}
\input{Sections/evalSetup-d}
\input{Sections/evalResults-d}
\input{Sections/related_work}
\input{Sections/discussion-d}
\input{Sections/conclusion}

\input{Sections/acknowledgements}}% if render_diff is false
{\input{Sections/introduction}
\input{Sections/motivation}
\input{Sections/method}
\input{Sections/evalSetup}
\input{Sections/evalResults}

\input{Sections/related_work}
\input{Sections/discussion}
\input{Sections/conclusion}}

\input{Sections/acknowledgements}
\bibliographystyle{ACM-Reference-Format}
\bibliography{lib.bib}
\input{Sections/appendix}

\end{document}

%% file: Sections/abstract.tex
\begin{abstract}
%\boldmath
%Speech, a prevalent human-computer interaction modality, is often transcribed into text, allowing us to use natural language processing techniques to extract its semantics. However, 

Dementia is a sensitive neurocognitive disorder affecting tens of millions of people worldwide and its cases are expected to triple by 2050. Alarmingly, recent advancements in dementia classification make it possible for adversaries to violate affected individuals' privacy and infer their sensitive condition from speech transcriptions. Existing obfuscation methods in text have never been applied for dementia and depend on the availability of large labeled datasets which are challenging to collect for sensitive medical attributes. In this work, we bridge this research gap and tackle the above issues by leveraging Large-Language-Models (LLMs) with diverse prompt designs (zero-shot, few-shot, and knowledge-based) to obfuscate dementia in speech transcripts. Our evaluation shows that LLMs are more effective dementia obfuscators compared to competing methods. However, they have billions of parameters which renders them hard to train, store and share, and they are also fragile suffering from hallucination, refusal and contradiction effects among others. To further mitigate these, we propose a novel method, \systemName{}. \systemName{} distills knowledge from LLMs using a teacher--student paradigm and parameter-efficient fine-tuning. \systemName{} has one order of magnitude fewer parameters compared to its teacher LLM and can be fine-tuned using three orders of magnitude less parameters compared to full fine-tuning. Our evaluation shows that compared to prior work \systemName{} retains the performance of LLMs achieving 1.3x and 2.2x improvement in privacy performance on two datasets, while humans rate it as better in preserving utility even when compared to state-of-the-art paraphrasing models.

\end{abstract}

%% file: Sections/related_work.tex
\section{Related Work}
\label{sec:rel_work}
\vspace{0pt}\noindent\textbf{Authorship and Attribute Detection.} Various stylometric features were identified to detect differences between authors such as lexical features, content-specific and syntactic features \cite{abbasi2008writeprints, schler2006effects, koppel2002automatically}. Several mental health conditions can also be detected from text~\cite{cohan-etal-2018-smhd,Edwards2017AMO,Wang2020LearningTD,bae2021schizophrenia}. While these works are crucial for early detection it is also a powerful tool for author profiling. Our work is the first to investigate dementia obfuscation in text.

\vspace{3pt}\noindent\textbf{Authorship and Attribute Obfuscation.} 
Authorship and attribute obfuscation techniques range from rule-based~\cite{castro2017author,karadzhov2017case} to paraphrasing and text generation. A4NT~\cite{shetty2018a4nt} and Style--pooling~\cite{mireshghallah2021style} perform obfuscation by style transfer by training a model to transfer sentences from one attribute to another or combining multiple styles. MutantX~\cite{mahmood2019girl} is optimized to perform word changes to fool the authorship detection model. Both in \cite{iyyer2018adversarial} and in \cite{mireshghallah2022mix}, authors learn syntactical changes to fool a classifier for sentiment, formality, and agency. Lin \& Wan \cite{lin-wan-2021-pushing} implement several iterations of back-translation in hopes of breaking the syntax and introducing synonyms at each pass. ParChoice~\cite{grondahl2020effective} implements a series of rule-based modifications for authorship and attribute obfuscation. It showed superior privacy performance while maintaining high semantic preservation compared to A4NT and MutantX. Differential privacy methods have also been applied for obfuscation in text, with most focus on authorship obfuscation.  Early work focuses on applying DP to the word level but often struggles to preserve semantics and computational efficiency. To counter these limitations, Mattern et al. \cite{mattern2022limits} suggest sampling from a language model using softmax scaled by a temperature.

\vspace{3pt}\noindent\textbf{LLMs Distillation.} Knowledge Distillation (KD) was introduced by Hinton et al.~\cite{hinton2015distilling}, to train a smaller model (student) to replicate the logit outputs of a larger, more complex model (teacher). %It was then followed by notable distillations work such as DistillBERT~\cite{sanh2019distilbert} or TinyBERT~\cite{jiao2019tinybert}, distilling the model at the intermediate hidden states level. 
With the emergence of LLMs and increased computational costs, KD techniques have been used to reduce models' sizes~\cite{abdin2024phi,team2024gemma,gu2023minillm} but also to create task-specific smaller-model that leverage the deeper knowledge of LLMs~\cite{hsieh2023distilling,jung2023impossible,xu2023inheritsumm}. Due to their size, these models are often distilled using ``black-box'' or sequence-level knowledge
distillation~\cite{kim2016sequence} approaches, only accessing the textual output for the student to train on.  Comparatively to our work, ~\cite{pham2023select,shleifer2020pre,calderon2023systematic,xu2023inheritsumm} and ~\cite{enis2024llm} performed sequence-level KD on the tasks of paraphrasing, text-simplification and neural machine translation. Text Laundering~\cite{jiang2023textlaund} employ a similar approach to mitigate adversarial attacks on downstream models. To our knowledge, we are the first to utilize KD from LLMs for attribute obfuscation in text.

%% file: Sections/conclusion.tex
\section{Conclusion}
\label{sec:conclusion}
In this work, we highlight the risk of leaking a sensitive medical user attribute from transcribed speech and leverage LLMs for privacy-preserving text transformations. We focused our analysis on dementia, a neurocognitive condition affecting tens of millions of people globally
and to the best of our knowledge, we are the first to attempt to obfuscate this very sensitive attribute in speech.
 First, we showed that transcriptions can already be accessed by a large proportion ($\sim$20\%) of the most popular mobile apps, a functionality often not properly described to users. Then, we investigated the capabilities of LLMs on the task of dementia obfuscation through various prompting strategies and found that zero-shot and domain-knowledge-informed prompts outperform compared to state-of-the-art (SOTA) text paraphrasing and attribute obfuscation models. Finally, we present \systemName{}, which distills knowledge from LLMs. \systemName{} is a significantly smaller student model compared to its teacher LLM model.
 Moreover, \systemName{} is partially fine--tuned for dementia obfuscation on a synthetic LLM-generated dataset with LoRA, which allows us to train and share three orders of magnitude fewer parameters compared to full fine-tuning.
 Our comprehensive evaluation demonstrated that \systemName{} retains the high obfuscation capabilities despite its smaller size and parameter-efficient fine-tuning. Our work highlights the capabilities of LLMs and knowledge distillation for attribute obfuscation in text and their potential for deploying effective and resource-efficient privacy-preserving text processing systems.

%% file: Sections/acknowledgements.tex
\begin{acks}
This study was funded by the Department of Computing at Imperial College London.
\end{acks}

%% file: Sections/introduction.tex
\section{Introduction}
\label{sec:intro}

\begin{figure}[!ht]
\centering
\includegraphics[width=1\columnwidth]{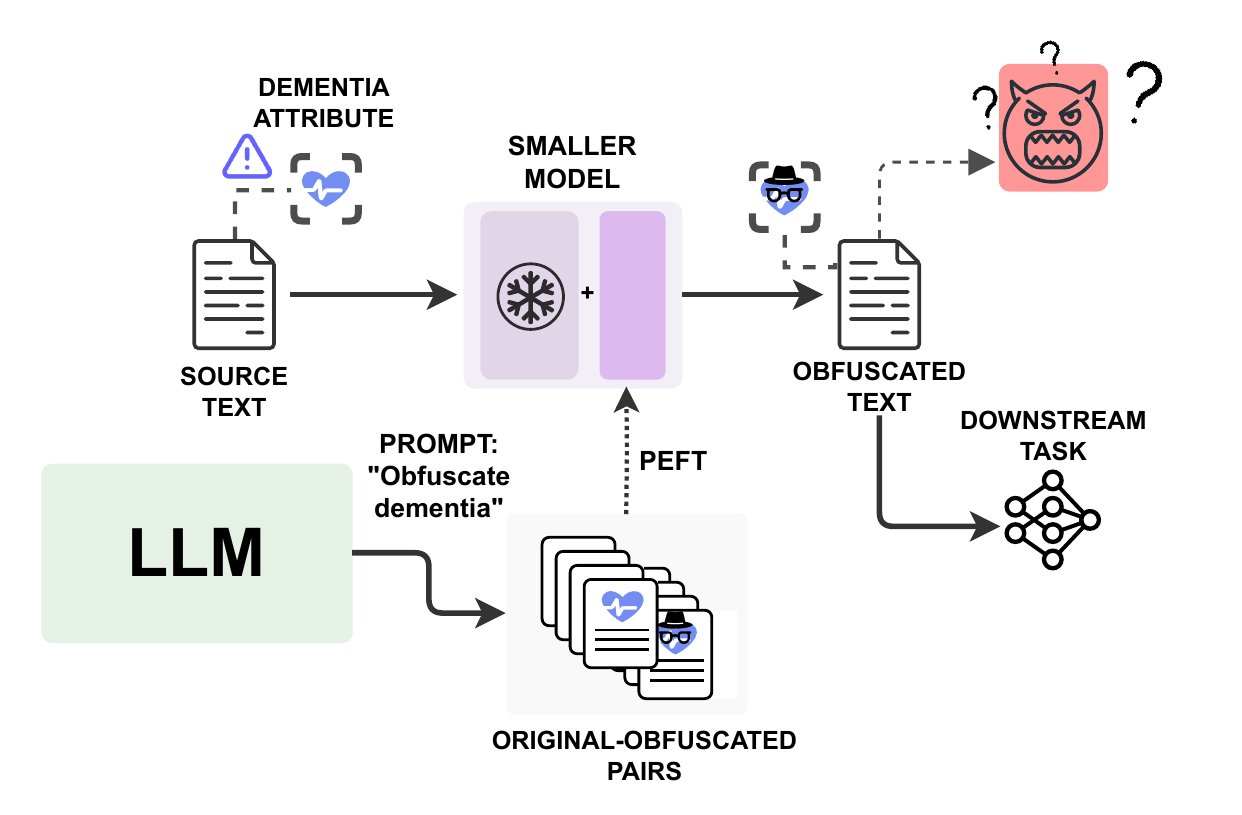}
\Description{Graph of the AD-GUARD framework. An LLM is used to generate a synthetic dataset of source-obfuscated sentence pairs for a smaller model to train on. The model is trained with PEFT. A source text with a dementia attribute goes through the framework which returns an obfuscated text. An adversary cannot detect dementia but the text can still be used for a downstream task.} 
\caption{Overall structure of \systemName{}. \systemName{} takes as an input a sentence with a potential dementia attribute and generates an obfuscated paraphrase.}
\label{fig:overall_sys}
\end{figure}

Speech is a prevalent human-computer interaction modality and there are strong business incentives for collecting and processing speech data. The size of the Voice Recognition and Speech processing global market amounted to \$12 billion in 2022 and is expected to grow to \$34 billion by 2028.~\cite{mordorintelligenceVoiceRecognition}. The voice assistant market alone is worth \$2.5 billion in 2023 and popular solutions such as Google Home and Amazon Alexa are being used by hundreds of millions of smart speaker users~\cite{statistaSpeakers}. Finance and Telecom companies increasingly rely on speech-to-text APIs and speech analytics to process calls for feedback, categorize queries, and check for compliance with regulations. The healthcare sector is the next biggest use case for speech-to-text as it can speed up note-taking and search and also analyse prevalent topics of worry and patient feedback~\cite{nhschoices}.
%alleviating the strain on the workforce and reducing waiting times for enquiries, and also analyzing prevalent topics of worry and patient feedback~\cite{}. 
Moreover, in this work, we found (see Section~\ref{sec:motivation}) that about 20\% of the most popular Android apps request permission to record audio out of which more than 79\% use APIs to process speech. Transcribing speech allows one to use natural language processing techniques to understand the semantics of the natural language. For example, \textit{Google Keep}~\cite{googleGoogleKeep} is a productivity app with more than 2 billion installations that allows users to take notes using their voice; \textit{Voice Recorder Audio Sound MP3}~\cite{googleVoiceRecorder} and \textit{Otter: Transcribe Voice Notes~\cite{googleOtterTranscribe}} have been installed more than 8 million and 1 million times respectively and produce live and offline voice transcriptions. 

\vspace{3pt}\noindent\textbf{Dementia Can be Inferred From Speech Transcripts.} 
Dementia, a degenerative neurological disorder, has become a pervasive global health concern with the ageing population. It affects individuals' cognitive abilities, memory, and overall functioning, and its cases are expected to triple by 2050~\cite{nichols2022estimation}. Disfluencies, mistakes, and signs of cognitive challenge are even more apparent in speech than they are in written text. Key discriminative features were found in speech transcripts and when aggregated, they can be important indicators of Alzheimer's Disease (AD). Recent works on the automatic detection of dementia, yielded promising results~\cite{yuan2020disfluencies,syed2021automated,zolnoori2023adscreen}. These, however, can also be exploited to extract sensitive medical information from transcribed speech. Personal information related to medical conditions is very sensitive and must be kept secret: employers can use it to discriminate against employees or applicants, banks can use it to associate loan applicants with higher risk scores, advertisers can use it to target users, etc. While user attribute leakage has been studied before~\cite{schler2006effects,sarawgi2011gender, emmery2018style, shetty2018a4nt}, little has been done against leakage of medical-related attributes from transcribed speech, such as dementia. 

This is worrisome and it might come in conflict with privacy regulations. For example, GDPR~\cite{gdpr} is an EU regulation that protects the processing of sensitive data. It defines sensitive data as personal data subject to several processing conditions including biometric data and health-related data~\cite{gdprSensitiveData}. GDPR also has \textit{data minimization} as one of its core data protection principles. According to data minimization, data processors must ensure that the data they process are adequate, relevant, and ``\textit{limited to what is necessary}''. Similarly,  
CPRA~\cite{cpra} (California Privacy Rights Act) which took effect in January 2023 and is an amended version of the CCPA (California Consumer Privacy Act) also stipulates that data processors should not process data beyond what ``\textit{reasonably necessary and proportionate to achieve the purposes for which the personal information was collected or processed}''. Therefore, there is a need for solutions that allow understanding of natural language from speech without allowing dementia inference.

%% PRIOR DEFENSE APPROACHES AND LIMITATIONS %%
%koppel2002automatically,abbasi2008writeprints,nguyen2013old
\vspace{3pt}\noindent\textbf{Prior Approaches.} 
Several works already exist aiming to either obfuscate sensitive attributes or identity~\cite{koppel2002automatically,abbasi2008writeprints,nguyen2013old,shetty2018a4nt,xu2019privacy, emmery2018style, mahmood2019girl, weggenmann2022dp, mattern2022limits}. Some follow a style-transfer approach to target gender, sentiment, age, and other attributes~\cite{shetty2018a4nt, emmery2018style, mahmood2019girl,xu2019privacy}. These \textbf{(a)} are limited by their dependence on the availability of training data for the specific attributes and \textbf{(b)} they require 
training a model for each new attribute and their obfuscation is hence classifier dependent. 
Medical datasets of conversations and written text for conditions such as dementia are hard to collect and access. Unlike existing style transfer datasets, parallel corpora between patients and control individuals are not available. This limitation calls for methods that can learn to perform obfuscation even in the absence of training data for the target attribute. 
Others have used differential privacy which can provide provable guarantees~\cite{weggenmann2022dp, mattern2022limits}. However, these suffer from poor semantic reconstruction. Furthermore, it is difficult to interpret the privacy metric's ($\epsilon$) value when applied to words in a sentence or a document. More importantly, none of the prior works have been applied to obfuscating sensitive medical attributes such as dementia.

\ignore{
Existing attribute and identity obfuscation approaches are inadequate. 
We can group prior relevant work broadly into three categories: (i) Authorship and attribute obfuscation in text~\cite{koppel2002automatically,abbasi2008writeprints,nguyen2013old,shetty2018a4nt}; (ii) style transfer approaches ~\cite{ shetty2018a4nt, emmery2018style, mahmood2019girl}; and (iii) differential-privacy-based approaches~\cite{weggenmann2022dp, mattern2022limits}. 
We can group prior relevant obfuscation work broadly into two categories:
(i) style-transfer approaches~\cite{shetty2018a4nt, emmery2018style, mahmood2019girl}; and (ii) differential-privacy-based approaches~\cite{weggenmann2022dp, mattern2022limits}.

Obfuscation in text and style transfer have been shown to be effective when applied to gender, sentiment, age, and other attributes, but  (a) they are limited by their dependence on the availability of training data for the specific attributes and (b) they require 
training a model for each new attribute and their obfuscation 
is hence classifier dependent ~\cite{ shetty2018a4nt, emmery2018style, mahmood2019girl}. Medical datasets of conversations and written text for conditions such as dementia are hard to collect and access. Unlike existing style transfer datasets, parallel corpora between patients and control individuals are not available. This limitation calls for methods that can learn to perform obfuscation even in the absence of large training datasets for the target attribute. More importantly, none of these have been applied for obfuscating medical attributes. Prior works using differential privacy have the advantage of providing provable guarantees. However, prior works~\cite{weggenmann2022dp, mattern2022limits} are concerned with authorship obfuscation and do not attempt to conceal specific attributes. More importantly, they suffer from poor semantic reconstruction as they increase the level of noise to conceal the authorship. Furthermore, it is difficult to interpret the privacy metric's ($\epsilon$) value when applied to words in a sentence or a document. 
}

\vspace{3pt}\noindent\textbf{Our Approach.} Our first key idea is to \textit{leverage the capabilities of large language models (LLMs) to perform zero-shot text generation and paraphrasing tasks}. We investigate several open-source LLMs that can be executed privately and design zero-shot, few-shot and knowledge-based instructions to significantly improve privacy performance compared to prior work. However, LLMs require costly memory and computational power due to their $\mathcal{O}(10^9)$ number of trainable parameters, making them impractical for many applications, and they can suffer from hallucinations and refusals. Our second key idea to overcome the LLM limitations is to perform \textit{teacher-student knowledge distillation from LLMs} in a 2-step process: first, we prompt a teacher LLM model to create a dataset of original and obfuscated sentence pairs. Second, we fine-tune a smaller, more efficient student model on this dataset. Fully fine-tuning the smaller model though, still requires training on $\mathcal{O}(10^8)$ parameters. To further improve on this we use our third key idea to \textit{apply a suitable Parameter-Efficient Fine-Tuning} (PEFT) technique. This involves freezing the student's layers, using matrix decomposition to reduce the number of parameters, and integrating lightweight adapter modules. This ensures efficient training, reduces overfitting risk on small datasets, and reduces computational overhead. Overall, our approach results in a smaller, efficient student model which we call \systemName{} (Knowledge \textbf{Di}stillation from Large-Language-Models for \textbf{D}ementia \textbf{O}bfuscation in \textbf{T}ranscribed \textbf{S}peech). \systemName{} retains the privacy and utility performance of the teacher LLM while requiring three orders of magnitude less parameters ($\mathcal{O}(10^5)$) for fine-tuning. Compared to a state-of-the-art attribute obfuscator (ParChoice~\cite{grondahl2020effective}), \systemName{} is not classifier dependent, achieves 1.3x and 2.2x better on average privacy performance across static and adaptive adversaries on two datasets, and a 3x improvement in a worst case scenario. Human evaluation shows that both our best LLM approach and \systemName{} achieved significantly better comparable utility performance than ParChoice and other baselines.

\ignore{
Our key insight is to use \hl{pairs of document paraphrases} to finetune an existing language model such that it can perform text generation with the intention of paraphrasing an original \hl{document.}
The model can be trained on a non-dementia dataset and applied directly to the sensitive dataset without further training. However, choosing the base model and the paraphrase pairs is not trivial as they determine the quality and structure of the output sequence. Adding to the challenge, text generation and paraphrasing models often produce sequences with non-standardized form, e.g. with grammatical errors, repetition, and nonsensical sentences.
This is problematic because as our analysis shows (see Section~\ref{sec:motivation}) such sequence characteristics are prevalent in dementia samples.
%In particular, we designed a set of features related to standardized speech encompassing fluency, and conciseness and introduced a new metric for analyzing dementia, formality. We then perform a test between dementia and control samples and found statistically significant differences between them for all dimensions.
To address this, firstly we chose BART~\cite{lewis-etal-2020-bart} as our base language model because it is shown to be more robust to errors and to generate more fluent sentences. 
Secondly, we used the PAR3 dataset~\cite{thai2022exploring}, a large-scale dataset of paragraph-level paraphrases, augmented with disfluencies to finetune the generic model for text paraphrasing. This allows us to train a model on longer sequences than current sentence-level paraphrase models and to steer the language model to removing disfluencies.

Another major challenge in obfuscating dementia is to preserve the semantics of the original sentence.
In Section~\ref{sec:motivation} we show that dementia classification can be very effective 
using lexical features. But perturbing lexical features has a penalty on semantics. However, we also show that syntax alone plays a major role in dementia classification. Perturbing syntax should have a much less important adverse effect on semantics.
%Moreover, syntax is independent of lexical features which are important for semantics.
 We leverage this insight and introduce syntactical diversity, using 
a conditional text generation approach.
%a \hl{prompt-tuning approach}~\cite{} 
We first teach the model to learn to generate text given two sentence pairs and their syntax distance
%first learn syntax distances between standardized sentence pairs 
and then prompt the text generation with a desired syntax control knob during inference.

We implemented the above methods into a novel end-to-end language model called 
\textit{Controlled Text Standardization} (CTS).
We evaluated the obfuscation ability of CTS against our most successful traditional and neural-based dementia classifiers. The adversarial classifiers are trained using real dementia datasets both in a static setting (with no knowledge of the defense mechanism) and a stronger adaptive setting where the adversarial models are trained on the output of CTS. We found that CTS is effective with adversaries  
achieving \textcolor{blue}{low F1 scores of 0.41 and 0.49 in the
best case.
We also found that CTS preserves semantics better than the best obfuscating system with Topic Modeling (0.72 vs 0.38 semantic similarity for CTS and BERTopic respectively) at comparable
obfuscation performance (mean of F1 scores of 0.39 for both)
and obfuscates better than a SOTA paraphrasing model for
similar semantic preservation values (0.74 SS for DP
with mean F1-scores of 0.54).}
}

\vspace{3pt}\noindent\textbf{Contributions.} Below we list our main contributions: 

\vspace{3pt}\noindent$\bullet$\textit{ New application domain.} We are the first to explore dementia obfuscation in transcribed speech. We hope this inspires further research on cognitive disorder obfuscation in speech.

\vspace{3pt}\noindent$\bullet$\textit{New Understanding.} We are the first to explore and rigorously evaluate the ability of LLMs to obfuscate dementia and preserve semantics. In doing so we have designed task-relevant zero-shot, few-shot and knowledge-based instructions for LLMs.

%\vspace{3pt}\noindent$\bullet$\textit{Novel Obfuscation Method.} We have designed a novel method that effectively distills knowledge from LLMs for dementia obfuscation in a parameter-efficient manner.

\vspace{3pt}\noindent$\bullet$\textit{Novel Obfuscation Method.} To overcome LLMs’ computational cost and brittleness (refusal, hallucination, contradictions), we have designed a unique and lightweight system that distills knowledge from LLMs for dementia obfuscation. Additionally, our approach leverages parameter-efficient training for a cost-effective and practical solution.

\vspace{3pt}\noindent$\bullet$\textit{ Rigorous empirical evaluation and findings.} We perform a systematic and rigorous evaluation of LLMs and our knowledge distillation method across both utility and privacy metrics and through ablation and human studies. We find that LLMs are better obfuscators compared to other competing approaches but have a very large number of trainable parameters ($\mathcal{O}(10^{9})$), and sometimes fail to follow instructions. Our novel \systemName{} method shows that we can learn to obfuscate similarly to LLMs while we only need to partially fine-tune a smaller model ($\mathcal{O}(10^{9})$) on three orders of magnitude less parameters ($\mathcal{O}(10^{5})$) compared to full fine-tuning.

%\vspace{3pt}\noindent$\bullet$\textcolor{blue}{\textit{ New insights.} Unlike previous studies on dementia features, we investigate a new set in view of spoken dementia obfuscation. Specifically, we select indicators relevant to standardization such as fluency, conciseness, and formality, and show significant differences between the two classes.}

%\vspace{3pt}\noindent$\bullet$\textit{ New approach.} We introduce an approach through text standardization that can be applied to low-resource datasets without specialized training which is suitable for obfuscating highly sensitive attributes with low dataset availability.

%\vspace{3pt}\noindent$\bullet$\textit{ New paraphrasing model.} We propose a novel obfuscation language model that can be applied to low-resource datasets. Unlike existing models, our model performs syntactically diverse and standardized sentence paraphrase generation which allows it to obfuscate dementia and preserve semantics.

\vspace{5pt}\noindent\textbf{Ethical Considerations.} In our work, we make use of available upon-request datasets ADReSS and ADReSSo from the TalkBank archive~\cite{macwhinney2004talkbank}, containing recordings and transcripts of conversations with dementia and control patients. Datasets submitted to the TalkBank conform to IRB practices i.e. they are anonymized and do not contain identifiable information. Our human subject study is approved from our institution's Research Ethics review board. Details of our study design are provided in Appendix~\ref{app:ethics}. 

%% file: Sections/motivation.tex
\section{Understanding the Risks}
\label{sec:motivation}

%\subsection{Security Threat Analysis}
%\subsection{Understanding the Risks}
%As individuals engage with a wide range of digital platforms, their interactions, and voices are increasingly being transcribed into text for further analysis. 

%To better understand the threats of profiling from transcribed speech we (a) 
%investigate the size, and nature of relevant markets, (b) perform a more detailed analysis on speech and transcription data collection and processing on the most popular smartphone OS ecosystem, and (c) identify alarming examples of mobile apps which process speech. 

%To better understand the threats of profiling from transcribed speech we perform a more detailed analysis of speech and transcription data collection and processing on the most popular smartphone OS ecosystem. 
%Then we define our threat model.

%, and (c) identify alarming examples of mobile apps which process speech. 
\ignore{
\vspace{5pt}\noindent\textbf{Market Overview.} The size of the Voice Recognition and Speech processing markets globally is\ignore{estimated to be around \$12 billion in 2022 and is} expected to grow to 34 billion by 2028~\cite{}. The speech-to-text API market and the speech analytics markets were valued at \$2.32 billion (2021)~\cite{} and \$2.3 billion (2022)~\cite{} respectively. Smart speakers also embody voice assistants (e.g. Amazon Alexa, Google Home etc) which collect and transcribe speech to aid in natural language understanding.  The voice assistant market is worth \$2.5 billion and hundreds of millions of users use smart speakers in their house~\cite{statistaSpeakers}. The top applications sectors were found to be healthcare,  banking, financial services and insurance (BFSI), retail, enterprise, consumer, automotive, and education for voice recognition~\cite{},  and the top use cases for Speech-to-text APIs were contact centres, transcriptions, fraud detection, compliance management, voice search and captioning~\cite{}.

%This is driven by several key players such as Google, Amazon, Microsoft, IBM, and Nuance Communications. 
%These technologies have found applications in diverse sectors, from healthcare and customer service to entertainment and education. The top applications sectors were found to be healthcare,  banking, financial services and insurance (BFSI), retail, enterprise, consumer, automotive, and education for voice recognition~\cite{},  and the top use case for Speech-to-text APIs were contact centers, transcriptions, fraud detection, compliance management, voice search and captioning.~\cite{}. Moreover, Amazon Alexa and Google Home have been leading the voice assistant market which is worth \$2.5 billion. More than \hl{X} million users make use of devices powered by Amazon Alexa and Google Home. As it is evident from the above, there are strong business incentives for collecting and processing speech data for millions of users across the world.

%(Shares of X and X respectively within the 2.5 billion total size)~\cite{} and popularising the use of voice-based interfaces among the population (X spread among US population).

%When surveying companies deploying voice solutions they found that “Customer Service” is the most deployed function (81\%), then “Sales” (52%), followed by “Store Operations” and “Marketing and Advertising ( both 38\%)~\cite{}.
}

%\vspace{5pt}\noindent\textbf{Speech-to-Text in Android apps.} To further quantify the prevalence of speech collection and processing, we conducted an analysis on popular mobile apps. 

\subsection{Speech--to--Text in Mobile Apps}
Existing market studies evidence the prevalence and importance of speech-to-text~\cite{enlyftWatsonSpeech,mordorintelligenceVoiceRecognition,statistaSpeakers,Speechto59:online}.
To better understand and highlight the threats of profiling from transcribed speech we perform a more detailed analysis of speech and transcription data collection and processing on the most popular smartphone OS ecosystem.
Specifically, we systematically quantify the prevalence of speech collection and processing, by analyzing popular mobile apps. We focus on mobile apps because they are a popular interface with digital services (more than 5 billion people own a smartphone~\cite{statistaSmartphoneUsers}) and because there are available tools for collecting and analyzing mobile apps.

Our analysis focuses on the Android ecosystem because it has the largest ($> 70\%$~\cite{androidMarketshare}) mobile OS market share. We collected the top $200$ free Android apps among $54$ application categories from GooglePlay~\cite{googlePlay}, Android's official application store, using \texttt{google-play-scraper}\footnote{https://github.com/facundoolano/google-play-scraper}. This resulted in $10070$ unique apps. For each app, we collected the Android package (\texttt{.apk} and application metadata such as the application's description, its requested permissions, and the number of installations. 

The first goal of our analysis is to study the prevalence of audio collection. Apps collecting audio could infer or leak sensitive information from speech either on purpose or inadvertently. The latter is possible if third-party libraries embedded in the app decide to take advantage of the legitimately granted permission to access audio and further process the audio to identify sensitive user attributes~\cite{grace2012unsafe}, or if the app is compromised~\cite{samsungKeyboardBug, demissie2020security, felt2011permission, tuncay2016draco}. To measure how many of these apps have the capability of collecting audio data, we search their \texttt{AndroidManifext.xml} files for the presence of the \texttt{RECORD\_AUDIO} permission. Developers use these files to provide important information (such as Android permissions) that the system requires to execute their apps. Alarmingly, we found that 20.3\%/10070 apps request permission. These apps have $128$M of installations on average with the most popular app having 15B installations.

Next, we wanted to further understand which of the
 apps which collect audio, specifically perform speech-to-text processing. To detect this, we wrote a script that looks for the presence of the system \texttt{SpeechRecognizer}~\cite{androidSpeechRecognizer} in the code of the apps. However, speech-to-text can be performed using third-party (3P) libraries as well. Therefore we further manually compile a list of 13 APIs based on market leaders~\cite{Speechto59:online} and popular services that we provided to our script. We supplement this list with the API's and popular libraries' SDK calls when available and keywords that search for functions that relate to speech-to-text (full list in Appendix~\ref{app:android_keywords}). Note, that our list is incomplete therefore our script will only provide a conservative estimate.
After removing apps for which decompilation failed, we were left with $1743$ apps.
We found that $1380$ apps ($79.1\%$) are currently employing speech-to-text. $1306$ ($74.9\%$) use the Android's built-in speech recognition system service \texttt{SpeechRecognizer}, 534 ($30\%$) matched for keywords associated with speech-to-text functions and 82 ($4.7\%$) use an API.
%These results indicate an active effort to transcribe or analyze spoken language. Looking at the category distribution we found that the majority of these apps are in the X categories.

To identify if any apps process speech in violation of users' expectations we investigated how many of the apps that use a speech-to-text API in their code do not explicitly mention it in their public descriptions~\cite{pandita2013whyper}. We randomly picked $50$ out of the $1380$ apps and manually went through their descriptions looking for mentions of speech-to-text or transcription activity. Specifically, we look for direct or implied (e.g. speak and it will repeat after you) mentions of ``speech-to-text'' or synonymous tasks such as transcription, dictation, live translation, captioning, and voice-based control. Alarmingly, we found that $42$ ($84\%$) of these apps do not mention it. After further analysis, we found that $1278/1306$ ($98\%$) use the \texttt{SpeechRecognizer} to enable voice search through Android's \texttt{SearchView} widget. Voice search allows the apps access to unfiltered transcriptions of users' queries~\cite{androidCreateSearch}.  This shows that voice control is a prevalent but not adequately described functionality which is increasingly offered by popular apps. To examine whether violations of user expectations happen not only for voice-controlled but also for other speech-to-text processing tasks, we
 repeated the experiment purposively sampling 50 apps that do not only use \texttt{SpeechRecognizer} (616/1380), therefore speech processing is unlikely to be triggered explicitly by the user. We found that 34/50 (68\%) do not refer to speech-to-text in their descriptions.

In summary, we found that (a) an important number of popular apps collect audio, (b) most of them transcribe speech, and (c) the transcription functionalities are rarely described in the apps' public descriptions.

%We noticed that  \texttt{SpeechRecognizer} is often linked to Android's \texttt{SearchView} widget which enables voice search and hence we repeated the experiment but limiting ourselves to the 616 apps that also had either keywords/SDKs or both. We found that only 16 out of 50 (32\%) refer to speech-to-text. In our analysis, we observe that unless speech is the core service or the app's description is very rigorous, the transcription or processing functionalities are often hidden.

\subsection{Preliminaries}
\label{sec:bkgd_nlp}

\vspace{5pt}\noindent\textbf{Text Feature Extraction.} 
To process text with algorithms and models, it must first be transformed into numerical form. Tokenization is the fundamental step, splitting words into sub-entities and mapping them to an index in a vocabulary. The simplest feature representation is the Bag-of-Words, which counts token occurrences in a document. The TFIDF matrix improves on this by scaling token frequency relative to the entire corpus, emphasising rare words (often nouns) with potentially more valuable information. However, these methods ignore grammar and word meaning. More advanced approaches like as Word2Vec\cite{mikolov2013efficient}, Glove\cite{pennington2014glove} or FastText\cite{bojanowski2017enriching} build word embeddings, mapping words into a vector space where similar tokens are close-by. Nevertheless, these embeddings lack context awareness, meaning the same word can have different meanings. Attention-based models like BERT~\cite{Devlin2019BERTPO} provide contextual representations, capturing semantic information for sentence level that can be fed to classification models such as neural networks. Contextual embeddings have been shown to hold powerful representations that can be then used for downstream tasks such as sentiment and semantic analysis. Structural information such as Part-of-Speech (POS) tagging and Constituency Parsing can also be extracted. POS tagging assigns grammatical tags to words, indicating their functions within sentences. Constituency parsing creates a syntax tree representing the hierarchical structure of sentences, which can be abstracted into grammar production rules defining permissible word arrangements. Both POS tags and syntax structure are valuable for tasks like attribute or authorship classification, machine translation, or named entity recognition.

\vspace{5pt}\noindent\textbf{Language Models.} A language model (LM) is a probabilistic model of a natural language. The model learns a conditional probability distribution $P(w)$ of the next word/token given the sequence of size $n$ of previous tokens. We can write the conditioned distribution as $P(w_{t_{n}}|w_{t_0},w_{t_1},...,w_{t_n-1})$.
Modern LMs, including Large Language Models (LLMs) like ChatGPT\footnote{https://chatgpt.com/} or Llama3\footnote{https://llama.meta.com/Llama3/}, are based on the Transformer architecture~\cite{vaswani2017attention} and predict text in an autoregressive manner (each predicted token depends on the previous ones). The Transformer consists of layers of self-attention mechanisms and feedforward neural networks. Self-attention helps encode the input contextually and capture long-range dependencies within a sentence, while cross-attention between encoded inputs and previous tokens aids in predicting the next token. The model then generates probabilities for each word in the vocabulary using a linear and softmax layer, at each step. Current LLMs are typically decoder-only models~\cite{brown2020language,abdin2024phi,team2024gemma}, using prompted text as input. 

Recent LMs are pre-trained on large corpora to predict or reconstruct masked sentences and words, learning the language's semantics and patterns in an unsupervised manner. These pre-trained models can then be used to extract meaningful representations useful for downstream tasks such as classification or measuring semantic similarity. The models can also be fine-tuned on specific tasks using smaller, labelled datasets, adapting them for tasks like question-answering or classification. However, with the advance of LLMs, carefully crafted prompts can achieve similar results without fine-tuning, as seen with models like GPT3~\cite{brown2020language}, InstructGPT~\cite{ouyang2022training} and FLAN models~\cite{chung2024scaling}. These models, trained on instructions and human feedback (reinforcement learning from human feedback), can perform tasks such as paraphrasing, text simplification, or style transfer in zero-shot and few-shot scenarios.

%\vspace{5pt}\noindent\textbf{Knowledge Distillation} 
%Knowledge or Model distillation involves transferring knowledge from a large, complex model to a smaller, more efficient one. This process often involves fine-tuning a smaller model on a dataset generated by the larger model. By distilling the knowledge learned by the large model into the smaller one, it's possible to create models that are computationally lighter while retaining much of the performance.

\vspace{5pt}\noindent\textbf{Dementia Classification.} Dementia is a condition that progressively deteriorates one's cognitive capabilities, affecting their ability to retain and recall information, reason and communicate. The most common form of dementia, Alzheimer's disease, accounts for approximately 60-80\% of cases. Early detection is key to preventing and providing care for patients with dementia. To this end, several studies investigated differences between a control group and patients with Alzheimer's and found significant changes in syntactical, lexical and acoustic indicators~\cite{fraser2016linguistic, le2011longitudinal,lancashire2009vocabulary}. For example, Le et al.~\cite{le2011longitudinal} perform a longitudinal study on four authors and have found that dementia patients demonstrate \emph{lexical indicators} such as a decline in vocabulary and in the usage of complex words, an increased number of low-specificity nouns and verbs, but also \emph{syntactical markers} such as lower frequency of nouns and higher frequency of verbs and lower usage of the passive form.
Recently, machine learning algorithms have been successfully applied for automatic dementia detection from speech (audio) and speech transcripts (text), presenting a cheaper, more efficient, and non-intrusive diagnosis method for patients. 
Garcia et al~\cite{de2020artificial} found that features extracted from the transcripts were often the most indicative of dementia. 
Others used a combination of linguistic and acoustic features, including \emph{pause} and \emph{disfluency} markers.
For example, Yuan et al.~\cite{yuan2020disfluencies} incorporated disfluency tags and silences in the transcript and achieved 89.6\% detection accuracy on the ADReSS \cite{luz2020alzheimer} dataset.

\subsection{Threat Model}
\label{sec:motivation:threat}
We consider any adversary with access to a target user's transcribed speech. The adversary's goal is to detect whether the target user suffers from dementia. The adversary can operate in two settings: (a) a static setting and (b) an adaptive setting. A static adversary ($\mathcal{A}$) is obfuscation-oblivious and has direct access to the raw text. $\mathcal{A}$ does not have access to any additional information such as age, mental state exam score or audio recordings and has to rely on the text or transcript only. The adversary can gain access to available upon request DementiaBank datasets, blogs or novels, as well as public knowledge of dementia-related stylistic properties. An adaptive adversary ($\mathcal{A}_{Ada}$) is obfuscation-aware. We assume that $\mathcal{A}_{Ada}$ has gained access to samples generated by any applied obfuscation strategy and can leverage these obfuscated samples to adapt and improve their detection capabilities. 

For both scenarios, we consider the adversaries able to implement a kernel-based text classifier (SVM) trained on word token frequencies and a neural network (BERT) able to learn more complex structures and contextual information within a sentence and documents. These were chosen from the ADReSS challenge. The SVM classifier is first described by Luz et al~\cite{luz2020alzheimer} in their paper describing and setting the baselines for the ADReSS Interspeech 2020 challenge. It achieved the best performance across several other classifiers on the test set when using only linguistic features. Our choice of SOTA is the BERT model which won the ADReSS challenge on detecting dementia from text~\cite{yuan2020disfluencies}.

%Our SVM baseline is taken from the paper introducing the ADReSS challenge~\cite{luz2020alzheimer} and our SOTA model from the winning work for that challenge on detecting dementia from text~\cite{yuan2020disfluencies}.

Our defense mechanism aims to perform automatic dementia attribute obfuscation from $\mathcal{A}$ and $\mathcal{A}_{Ada}$ in text while preserving the semantics of the original transcribed speech.

%% file: Sections/method.tex
\section{Dementia Obfuscation in Text}

\begin{figure*}[!h]
\centering
\includegraphics[width=0.9\textwidth]{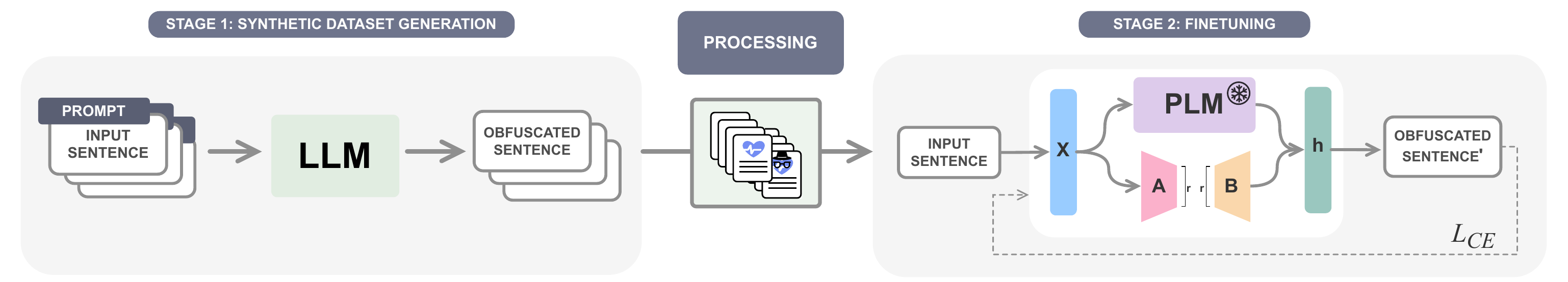}
\Description{Two sequential blocks showing that an LLM is prompted to generate an obfuscated sentence for each sentence in the set. This dataset goes through processing and is used as input to the second block, model fine-tuning. The figure shows a frozen pre-trained language model fine-tuned with LoRA layers.}
\caption{The two steps of \systemName{}: 1) the creation of a synthetic obfuscated dataset and 2) finetuning a smaller pre-trained language model (PLM) on this dataset with LoRA and a reconstruction loss $L_{CE}$ (cross-entropy). The weights of the PLM are frozen and only matrices A, and B are finetuned. At inference time, the PLM weights are combined with the learned matrices and the model performs obfuscation on input sentences.}
\label{fig:detailed_sys}
\end{figure*}

In this work, we present an approach that takes advantage of the knowledge and rich representations of large-language models (LLMs) to perform dementia obfuscation in text. However, LLMs come with limitations and constraints, the most important one being their size and their operational overhead. Indeed, they require a large amount of resources to both hold them in memory and perform inference in a reasonable time. To tackle this issue, we perform knowledge distillation by prompting an LLM to perform obfuscation (stage I) and recording the [\textit{original}, \textit{obfuscated}] pair to train a smaller and more efficient model on a synthetic dataset (stage II). We further enhance the training efficiency and modularity of our approach fine-tuning our model through a Parameter-Efficient Fine-Tuning (PEFT) method. We describe the different stages of (shown in Figure~\ref{fig:detailed_sys}.) of our methodology in the sections below.

%\subsection{LLMS for AD obfuscation}

\subsection{Knowledge Distillation}

Dementia obfuscation is particularly challenging as sparse data and no parallel datasets are available. This situation is similar to unsupervised style transfer, where various methods have been explored. 
A naive solution is to define a set of transformations. However, such rule-based systems~\cite{karadzhov2017case, castro2017author}, require manual design, lack the flexibility of trained models and may fail to capture the full range of linguistic patterns associated with dementia. Back-translation has been used for style transfer, involving translation to another language and back to the original language~\cite{prabhumoye2018style,lin-wan-2021-pushing}. To enforce a target style, Prabhumoye et al.~\cite{prabhumoye2018style} investigate back-translation combined with adversarial learning and cycle consistency. Similarly, Shetty et al. and Iyyer et al.~\cite{shetty2018a4nt, iyyer2018adversarial} employ a GAN-like model to perform style transfer. However, these methods risk overfitting on small datasets, struggle with semantic reconstruction, and are difficult to train due to multiple balancing losses. Imposing a semantic reconstruction loss on an auto-regressive model is non-trivial (the loss is computed on one token at a time) and requires the use of more complex systems such as reinforcement learning~\cite{phatak2022medical, xu2016optimizing,luo2019dual} or human-in-the-loop~\cite{ghodratnama2021towards}, adding complexity, instability and manual labour (in the case of human-in-the-loop). 

Controlled text generation is an alternative approach which guides the model during inference, requiring less data and computation~\cite{kumar2021controlled, Keskar2019CTRLAC,mireshghallah2022mix}. However, it still depends on using a target attribute classifier and the base model's capabilities, which may be insufficient in smaller models, leading to challenges in fluency and semantic reconstruction. One might try to focus on controlling syntax and linguistic features only, releasing the dependency on the dataset. Nevertheless, these approaches introduce multiple fine-grained control knobs and struggle to generate quality sentences~\cite{bandel-etal-2022-quality,iyyer2018adversarial}. Enforcing a specific style or lack of it without a classifier is also challenging in that setting.

LLMs on the other hand, are powerful foundational models which hold rich representations and achieve state-of-the-art performance on many zero-shot tasks. Hence, LLMs offer a more direct solution for scenarios with limited, non-parallel data. However, their immense size (Gemini Ultra by Google has 1.75 trillion parameters ($\mathcal{O}(10^{12})$~\cite{team2024gemma}) makes them costly to train and inaccessible to most institutions. Additionally, LLMs may exhibit hallucinations or refuse to follow instructions, rendering them less reliable. Model compression techniques like quantization~\cite{team2024gemma} and pruning~\cite{mehta2024openelm,Llama3is11:online} can reduce LLM size, but these methods often result in accuracy loss. Quantization, in particular, is a lossy process that can degrade model performance~\cite{schaefer2023hardware}.

Given these challenges, knowledge distillation (KD) on LLMs emerges are the most efficient and straightforward strategy. Our key insight to overcome the limitations imposed by a small, non-parallel dataset while taking advantage of the powerful nature of LLM in text generation is to \textit{distill knowledge from large language models into a smaller model that is easier to train, store and share, and generates more stable output}.

\vspace{3pt}\noindent\textbf{Teacher-student Model.} In our work, we leverage KD introduced by Hinton et al.~\cite{hinton2015distilling}, where we transfer knowledge from a large, complex model (teacher) to a smaller, more efficient one (student). Specifically, we perform sequence-level knowledge distillation~\cite{kim2016sequence} by generating a synthetic dataset using the larger model's outputs. This approach is more efficient than approximating the teacher's internal states due to the LLMs' size. By distilling the knowledge already learned by the larger model into the smaller one, we create a model that is computationally lighter while retaining much of the performance of the original model. This approach avoids sacrificing the accuracy or complexity of the resulting model, in favor of a bias toward examples specific to our downstream task (fine-tuning).

%\vspace{3pt}\noindent\textbf{Teacher Model} 
Our method is illustrated in Figure~\ref{fig:detailed_sys}. In the first stage, we use an LLM as a teacher model to generate [\textit{original}, \textit{obfuscated}] sentence pairs to create a synthetic dataset to, in stage 2, teach a smaller pre-trained language model (PLM).
Our student model implementation is based on a pre-trained BART model~\cite{lewis-etal-2020-bart}, which is built as a classical Auto-Encoder Transformer. We chose BART due to its good performance and generation quality. Thanks to its denoising pre-training task, BART is more robust to input noise and generates fluent sentences. This is desirable because dementia sentences are less formal and other paraphrasing models tend to produce sentences in non-standardized form. The model is fine-tuned on the synthetic dataset through supervised fine-tuning with cross entropy-loss. Note that BART can easily be replaced in this process with another sequence--to--sequence model. 
%In our evaluation (Section~\ref{sec:evalResults}) we demonstrate the effect of the student model choice.

%\subsection{Data Labeling \& Prompting Strategies}

\begin{figure}[!h]
\centering
\includegraphics[width=\columnwidth]{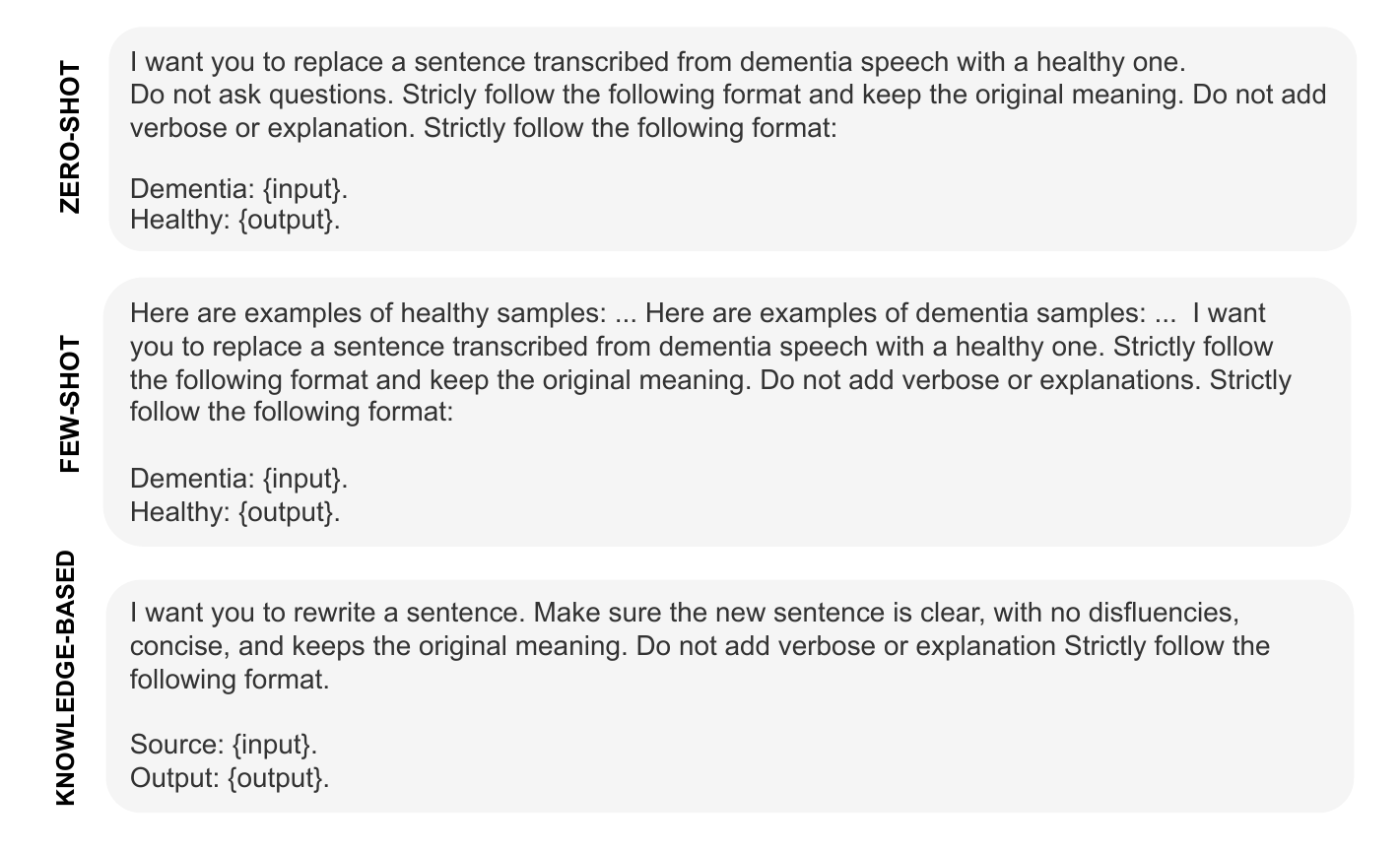}
\Description{Three bubbles with the text prompt used to query the LLMs: zero-shot, few-shot, knowledge-guided}
\caption{Instructions for zero-shot, few-shot and knowledge-based prompting strategies. We omit few-shot examples for the sake of space.}
\label{fig:prompt_strats}
\end{figure}

\vspace{3pt}\noindent\textbf{Data Labeling \& Prompting Strategies.} 
%Our approach embodies a stage where a synthetic dataset is generated using instruction-based LLMs.  
A key step of our approach is synthetic data generation by prompting instruction-based LLMs and relies on the quality of instructions provided to the selected LLM. We follow and compare several strategies to achieve good-quality outputs. Specifically, we prompt an instruction-based LLM to generate an obfuscation candidate for every sentence in our dataset. We use three different prompting strategies as shown in Figure~\ref{fig:prompt_strats}: \textit{zero-shot}, \textit{few-shot}, and \textit{knowledge-based}. To all prompts, we add safeguard instructions to force the model to follow the prompt as closely as possible and avoid rejections or noisy answers.  
In the zero-shot setting, we ask the model to obfuscate a dementia sentence. In the few-shot setting, the model is given 10 examples of dementia and control sentences. Finally, in the knowledge-based prompt, we incorporate knowledge from previous works and dementia detection studies. These have identified that dementia patients tend to be more wordy, and vague, use simpler vocabulary and display highly disfluent speech~\cite{fraser2016linguistic, lancashire2009vocabulary, le2011longitudinal, becker1994natural}.
Based on this we construct an instruction for the LLM asking it to clarify a given sentence. Moreover, in our instruction, we do not mention dementia. This strategy is employed to circumvent the censorship showcased by some LLMs when sensitive topics are mentioned. In the optic of ensuring `alignment' (a term designating the capacity of a model to follow given ethical and legal guidelines) and reducing risks linked to chatbot usage, companies such as OpenAI, Meta or Google implement various safety measures. These can include experts red teaming and evaluating the safety of prompts~\cite{meta:online, OpenAIRe94:online} and filtering data used for pre-training or post-generation moderation~\cite{openai_mod:online,team2024gemma,abdin2024phi,touvron2023llama}. 
%These policies are put in place to prevent harmful content such as hate speech or promote illegal activities. 
Reportedly, ChatGPT and Llama2 models have shown high levels of censorship (on topics related to politics, gambling or ``killing'' time)~\cite{reuter2023m,Llama3is11:online,Ididntth65:online}, often misinterpreting a benign prompt as unsafe (see examples in Table~\ref{tab:hits_miss}). After generation, samples are cleaned from the template to obtain the obfuscated sentence.

\subsection{Parameter-Efficient Fine-Tuning}
As mentioned, we fine-tune our student model using the synthetic dataset. However, we do not perform full fine-tuning but instead employ a Parameter-Efficient Fine-Tuning (PEFT) technique. PEFT is a new paradigm that involves optimizing pre-trained language models for specific tasks by fine-tuning only a small subset of parameters~\cite{hu2021lora, liu2021p, lester2021power}. Through PEFT we aim to retain the majority of the student model's pre-trained knowledge, making it resource-efficient and faster than full model fine-tuning. Also, by focusing on fewer trainable parameters, we further reduce the risk of overfitting and maintain the model's overall performance.

More specifically we fine-tune our student model using Low-Rank Adaptation (LoRA)~\cite{hu2021lora}, a state-of-the-art PEFT technique. With LoRA, we inject trainable low-rank matrices into each layer of the Transformer architecture, allowing only these matrices to be fine-tuned while the rest of the model remains frozen. Therefore, we can drastically reduce the number of trainable parameters, making the fine-tuning process efficient and preserving the LLM's core knowledge.

With LoRA, the adaptation of the model is achieved by decomposing the weight updates into low-rank matrices. Specifically, for a given weight matrix $W$ in the transformer model, we introduce two smaller matrices $A$ and $B$ such that:

\[
\Delta W = A \cdot B
\]

where $A \in \mathbb{R}^{d \times r}$ and $B \in \mathbb{R}^{r \times k}$, with $r \ll \min(d, k)$. Here, $d$ and $k$ are the dimensions of the original weight matrix $W$. During fine-tuning, only the parameters in $A$ and $B$ are updated, keeping the original $W$ unchanged. The effective weight matrix $W'$ during adaptation thus becomes:

\[
W' = W + \Delta W = W + A \cdot B
\]

This low-rank decomposition significantly reduces the number of parameters that need to be trained, as the number of parameters in $A$ and $B$ combined is much smaller than in the original $W$.

%During training, only the parameters associated with LoRA Modules are updated.

%% file: Sections/evalSetup.tex
\section{Evaluation Setup}
\label{sec:evalSetup}

\subsection{Evaluation Questions}\label{subsec:evalQuestions}
We conduct a comprehensive evaluation aiming to answer the following concrete questions: (\textbf{EQ1}) How effective is our knowledge distillation method through an LLM--generated dataset?;
(\textbf{EQ2}) How do our proposed obfuscation methods compare to state--of--the--art and baseline approaches?;
(\textbf{EQ3}) How effective are LLMS in obfuscating dementia and preserving semantics?;
(\textbf{EQ4}) How do different components of our knowledge distillation method contribute to performance (utility/privacy)?;
(\textbf{EQ5}) How do obfuscation systems impact key dementia discriminative features (privacy)?;
(\textbf{EQ7}) How do the generated sentences compare qualitatively with the original sentences?

\begin{comment}
\vspace{3pt}\noindent$\bullet$\textbf{ EQ1.} How effective is our knowledge distillation method through an LLM--generated dataset?

\vspace{3pt}\noindent$\bullet$\textbf{ EQ2.} How do our proposed obfuscation methods compare to state--of--the--art and baseline approaches?

\vspace{3pt}\noindent$\bullet$\textbf{ EQ3.} How effective are LLMS in obfuscating dementia and preserving semantics?

\vspace{3pt}\noindent$\bullet$\textbf{ EQ4.} How do different components of our knowledge distillation method contribute to performance (utility/privacy)?

\vspace{3pt}\noindent$\bullet$\textbf{ EQ5.} How do obfuscation systems impact key dementia discriminative features (privacy)?

\vspace{3pt}\noindent$\bullet$\textbf{ EQ6.} What is users' perception of our methods' ability to preserve semantics (utility) and how does it compare with other methods?

\vspace{3pt}\noindent$\bullet$\textbf{ EQ7.} How do the generated sentences compare qualitatively with the original sentences?
\end{comment}

\subsection{Datasets}
%\vspace{5pt}\noindent\textbf{Dementia Datasets.}  
We use two available datasets which are widely studied in the field of dementia classification: ADReSS~\cite{luz2020alzheimer} and ADReSSo~\cite{luz2021detecting}. The ADReSS (ADR) dataset is a curated subset of the DementiaBank dataset~\cite{becker1994natural}, a collection of 500 recordings from individuals with various stages of dementia. The samples were manually transcribed with dysfluency annotations for the task of the Cookie Theft Picture description. The speakers were selected to be balanced for gender and age, and they split evenly into control (CC) and dementia (AD) groups (54 train and 24 test for each class, a total of 156 samples). In our experiments, we use the provided train/test splits and split the documents on sentence level. We further remove sentences with less than 3 words. We end up with 1179 samples (619 CC $|$ 560 AD) in the training set and 500 (270 CC $|$ 230 AD) in the test and validation sets. The ADReSSo (ADRo) dataset is another subset of DementiaBank designed for detecting dementia from spontaneous speech only, without access to manual transcriptions. The original set consists of 151 train samples (87 CC $|$ 74 AD) and 71 test samples (35 CC $|$ 35 AD). We use the provided segmentation timestamps to isolate segments spoken by the patients and transcribe them using Whisper, a SOTA speech-to-text model~\cite{radford2023robust}. We get 1421 samples in the train set (710 CC $|$ 711 AD) and 644 samples (347 CC $|$ 297 AD) in the test and validation sets. 

\subsection{Adversarial Models}
In evaluating the ability of obfuscators to preserve privacy it is important to consider an adversary in both a static and an adaptive setting. In the former case, the adversary uses available datasets to train the adversarial model. In the second, stronger scenario, the adversary also has knowledge of the defense mechanism.
We implement both a static and adaptive adversary as described below.

\vspace{3pt}\noindent\textbf{Static Adversary.} We finetune a pre-trained BERT-base model with a classifier head as our neural adversary. BERT has been shown to have SOTA performance for various NLP tasks ~\cite{fabien-etal-2020-bertaa,sun-etal-2019-utilizing,bert_age_gender}. We use the base model from\cite{yuan2020disfluencies}\footnote{We evaluate both the base and large versions but pick the base model as it was more robust to obfuscation.} with a learning rate of 1e-6, 10 epochs with early stopping (patience =1 on validation loss), gradient clipping of 1, input length limit of 256 tokens,  and batch size of 8. We also select an SVM as a kernel-based baseline, as presented by ~\cite{luz2020alzheimer} and ~\cite{woszczyk22_interspeech}. We extract a TF-IDF matrix from the corpus and feed it to an SVM with radial-basis-function kernel and $C$ =1.0. The TF-IDF matrix also focuses on different features with learned word frequencies and relative frequencies while BERT looks at the big picture and overall fluency.

\vspace{3pt}\noindent\textbf{Adaptive Adversary.} In addition to the static adversaries we also evaluate our systems against adaptive adversaries with knowledge of the defense mechanism. We assume the adversary knows which obfuscator is being used, and that they have black-box query access to it. We train both SVM and BERT models with training data which includes raw data and their obfuscated version for a given defense.

%Topic Modeling (no syntax), Pegasus Paraphrase with T (syntax + semantics), ParChoice (rule-based), Mutant-X, A4NT, Pick most diverse syntax + Best model to reproduce(AESOP-h4?)/ SCPN?

%Our Models: prompt-base syntax BART, syntax-bart with code embeddings.

\subsection{Metrics}
\vspace{3pt}\noindent\textbf{Privacy.} We measure the privacy gain through the drop on the various adversaries' F1-score, for both static and adaptive settings. We refer to the F1-score as the adversary success rate (ASR). We use the F1 metric as it is less sensitive to data imbalance, and shows the model's ability to capture dementia cases and transform dementia samples to control-like ones. 

\vspace{3pt}\noindent\textbf{Utility.} There are currently no available dementia datasets that are labeled for specific tasks such as topic modeling or sentiment analysis. Therefore, for our utility metric, we
focus on semantic preservation. Semantic preservation in natural language is
a powerful primitive that is important for several downstream tasks such as topic modeling, patient and customer complaint analysis, note-taking, etc. 
%To evaluate semantic preservation we conduct (a) comparison with automated metrics, (b) comparison with human subjects, and (c) qualitative comparison of output sentences. 
 We automatically compute the semantic similarity between the original sentence and the output of a selected obfuscation model. We use ParaBART~\cite{huang2021disentangling}, a BART-based model trained to create syntax-invariant semantic embeddings. Common semantic embedding models suffer from poor robustness to different syntax and thus would lower the semantic score between two sentences with different structures even if the meaning is the same. ParaBART on the other has been shown to be more robust to such changes. 

Automated metrics, however, do not always align with human judgment. To mitigate this threat to the validity of our evaluation, we also design a targeted study with human subjects. Our study is designed to measure users' perception of the ability of different text paraphrasing models to preserve the semantics of original sentences. Our study was reviewed and received approval from our institution's research ethics review board. Lastly, we further provide a qualitative analysis of the paraphrase quality of different systems: we select representative sentences from our dataset and compare them with the output of competing systems.

%The benefit of automated metrics is that they allow us to scale our analysis to several systems and parameters as opposed to human subject studies and qualitative analyses which are conducted on a sentence by sentence case. In addition human studies are cumbersome to design and conduct, time-consuming and costly. 

\vspace{3pt}\noindent\textbf{Quality and Diversity Metrics.} To better understand the effect of different systems we measure the quality and diversity of the generated samples by computing the formality score, METEOR score (a common metric that measures word matches, synonyms, and word order.)~\cite{lavie2009meteor}, lexical similarity (Simi$_{Lex}$) measured by the word-level Levensthein distance between sets of unique words, perplexity and the percentage of substitutions (\% SUB), additions (\% ADD) and deletions (\% DEL) between the source and generated samples. We compute an automatically generated formality score of a sentence by using a RoBERTa model from the Hugging Face library\footnote{https://huggingface.co/s-nlp/roberta-base-formality-ranker} trained on the formality-style transfer dataset GYAFC ~\cite{rao-tetreault-2018-dear} and online formality corpus~\cite{pavlick2016empirical}. Given that neural text generation creates samples with word distribution that are distinct from human-written text, we also investigate the perplexity (PPL) of our generated samples. A high perplexity is commonly associated with human text but is also a sign of low fluency. 
%We use this metric to monitor fluency loss or gain between different approaches. 
We compute the perplexity by computing the exponentiated average negative log-likelihood of a pre-trained GPT2 model using the \textit{lmppl} library\footnote{https://github.com/asahi417/lmppl}.

%Apart from our utility and privacy metrics, text generation methods need to also demonstrate ability to generate high quality and diverse sentences.  We measure the quality and diversity of the samples generated by different LLMs leveraging established text metrics such as the length of the newly generated samples, \textit{lexical diversity}, the \textit{formality} score, and \textit{perplexity} (described above). The formality score is computed as described in Section~\ref{sec:motivation}. In addition, we also measure the percentage of refusals (no paraphrase returned by the LLM) as this is an important limitation of LLM usage.

\subsection{Obfuscation Models}

    \vspace{3pt}\noindent\textbf{Pegasus (Paraphrasing Model).} We compare our approach to the state-of-the-art paraphrasing model. We employ a Pegasus model~\cite{zhang2020pegasus} fine-tuned to perform paraphrasing on the input text. Pegasus is a transformer-based model that has shown impressive performance in generating high-quality text and paraphrases. It was pre-trained on a large corpus with the task of retrieving important masked sentences and later fine-tuned on paraphrase pairs.
    
    \vspace{3pt}\noindent\textbf{DP (Differential Privacy Obfuscation Model).} 
    Following work by~\cite{mattern2022limits}, we re-implement a text generation model that performs obfuscation by sampling with higher temperatures. We select Pegasus as our language model and do our best to improve on~\cite{mattern2022limits} ---our version is bigger and trained on a larger dataset than their model, to maximize the quality of the generated samples. We also experiment with various temperature values (see Appendix~\ref{app:dp}).

    \vspace{3pt}\noindent\textbf{ParChoice (Heuristics-based Obfuscation Model).} ParChoice is a rule-based obfuscation model that has been shown to have good obfuscation abilities while preserving semantics and utility~\cite{grondahl2020effective}. The model performs a series of modifications such as changing writing voice, contractions, typos, synonyms, etc. It also performs paraphrase selection via a surrogate profiler for style transfer and filter-generated sentences for fluency and grammatical correctness. This approach outperforms simpler rule-based approaches, but also complex models like A4NT~\cite{shetty2018a4nt} and MutantX~\cite{mahmood2019girl} for both obfuscation and semantic retainment.  We use the code made available by the authors and run ParChoice in ``Random'' settings as well as with an SVM and BERT surrogates trained on ADR samples that guide the obfuscation toward a control-like style.

    \vspace{3pt}\noindent\textbf{LLMs (Ours)} We investigate top-performing LLMs for different size brackets in their ability to perform dementia obfuscation.  Due to the sensitive nature of the data, we only use open-sourced models that can be run locally. As ChatGPT and other OpenAI models access and process input data, we omit them in this study. We select Gemma 2B~\cite{team2024gemma}, Phi3 (3B parameters)~\cite{abdin2024phi}, Mistral 7B~\cite{jiang2023mistral} and LLama3 8B~\cite{Llama3is11:online}. We pick Mistral as our best system to train \systemName{} and present all three prompt settings in the summary Table~\ref{tab:overall-results}. We run each model via the Ollama python library~\footnote{https://github.com/ollama/ollama-python}.
    
\vspace{3pt}\noindent\textbf{\systemName{} (Ours)} We implement our model using the pre-trained \textit{BART-base} model from the Huggingface library\footnote{https://huggingface.co/facebook/bart-base}. We implement and train the core of our system with LoRA (r=16, alpha = 32) on the KB dataset: \systemName{}, and our ablation models: BART$_{KB}$ (full fine-tuning), T5$_{KB}$ (T5 model\footnote{https://huggingface.co/google-t5/t5-base}), BART$_{ParaNMT}$ (trained on paraphrasing dataset ParaNMT~\cite{wieting2017paranmt}), as well as BART$_{KB\_BOFT}$ and BART$_{KB\_IA3}$ for different PEFT techniques: IA3 and BOFT. Intrinsically-Aligned Adapters (IA3) is an adapter technique that injects a small number of trainable parameters (64.5K) into the model. Bottleneck Output-Fusion Transformers (BOFT) introduce a bottleneck layer (138K) to compress and then expand the intermediate representations, to focus on relevant features. We use the paper's default parameters for both.
We finetune our models on the synthetic paraphrase pairs for a maximum of 3 epochs and use early stopping (patience = 10 on validation loss), with a batch size of 8 and a learning rate of 1e-4. At inference time, we sample with the number of beams set to 4 and a maximum length of 256.

%% file: Sections/evalResults.tex
\section{Evaluation Results}
\label{sec:evalResults}

\subsection{Overall Performance and Comparison}\label{sec:evalSetup_overall}
In this section, we aim to answer EQ1 and EQ2 (see Section~\ref{subsec:evalQuestions}).
We compare several baselines (\textit{Pegasus}) and SOTA (\textit{DP} and \textit{ParChoice}) models with our best performing LLM (Mistral) used with zero-shot (ZS), few-shot (FS) and knowledge-based instructions (KB), and with our knowledge distillation approach from Mistral with LoRA adaptation (\systemName{}). Note that a more comprehensive evaluation of LLM performance is presented in Section~\ref{subsec:llmEval}. 

The systems are evaluated for privacy against both our static and adaptive adversary and for utility (semantic preservation). We also indicate the size of each model when relevant as one of our main objectives is to distill knowledge from large language models, into a smaller model. Lastly, we also evaluate the above on both the ADR and the ADRo datasets. Evaluating on ADRo allows us to examine the effects of leveraging automated speech recognition to derive transcriptions for the adversary and the defense mechanisms and is a more challenging scenario compared to the manually transcribed samples of ADR. We summarize the results in Table~\ref{tab:overall-results}.

%\textcolor{red}{Why is it important to show performance on ADReSSo? In what settings this might be useful or more realistic compared to ADReSS?} T

\begin{table*}[!ht]
\centering
\caption{Privacy/utility performance for ADReSS and ADReSSo datasets by different systems. This table presents the F1 score against static (S) and adaptive (A) dementia classifiers, and the average F1 score (Avg F1) across all classifiers. It also depicts semantic preservation (Sem.). We mark our best systems in bold. $\downarrow$ illustrates the defense mechanism aims to decrease that value, and $\uparrow$ to increase it.}
\label{tab:overall-results}
\resizebox{\textwidth}{!}{
\begin{tabular}{lll|cc|cc|>{\columncolor{lightblue!35}}c|>{\columncolor{lightblue!35}}c|cc|cc|>{\columncolor{lightblue!35}}c|>{\columncolor{lightblue!35}}c}
\toprule

\multirow{3}{*}{\textbf{Type}} &               \multirow{3}{*}{\textbf{Systems}}       &           \multirow{3}{*}{\textbf{\# Params}}           & \multicolumn{6}{c|}{\textbf{ADReSS}}  & \multicolumn{6}{c}{\textbf{ADReSSo}}    \\
\cmidrule(lr){4-9} \cmidrule(lr){10-15}
              &               &             & \multicolumn{2}{c|}{\textbf{BERT$\downarrow$}} & \multicolumn{2}{c|}{\textbf{SVM$\downarrow$}} & {\cellcolor{white}}\multirow{2}{*}{\textbf{Avg F1$\downarrow$}} & {\cellcolor{white}}\multirow{2}{*}{\textbf{Sem.$\uparrow$}}  & \multicolumn{2}{c|}{\textbf{BERT$\downarrow$}} & \multicolumn{2}{c|}{\textbf{SVM$\downarrow$}} & {\cellcolor{white}}\multirow{2}{*}{\textbf{Avg F1$\downarrow$}} & {\cellcolor{white}}\multirow{2}{*}{\textbf{Sem.$\uparrow$}}  \\
\cmidrule(lr){4-5} \cmidrule(lr){6-7} \cmidrule(lr){10-11} \cmidrule(lr){12-13}
                  &                      &                      & \textbf{S} & \textbf{A} & \textbf{S} & \textbf{A} & {\cellcolor{white}} \textbf{}  & {\cellcolor{white}} \textbf{} & \textbf{S} & \textbf{A} & \textbf{S} & \textbf{A} & {\cellcolor{white}} \textbf{} & {\cellcolor{white}} \textbf{} \\
\midrule
      Original            & -             & -                    & 0.65        & 0.65          & 0.65        & 0.65          & 0.65 & 1.0 & 0.66       & 0.66         & 0.63       & 0.63         & 0.65 & 1.0            \\
\midrule
    Baselines              & Pegasus              & 568M                 & 0.52       & 0.58         & 0.60       & 0.59         & 0.57 & \textbf{0.85} & 0.5        & 0.57         & 0.58       & 0.58         & 0.56 & \textbf{0.81}       \\
                  & DP$_{T = 1.5}$     & 568M                 & 0.51       & 0.6          & 0.59       & 0.56         & 0.57 & 0.84 & 0.47       & 0.63         & 0.58       & 0.61         & 0.57 & 0.77       \\
                  \cmidrule(lr){2-15}
                  & Parchoice$_{Random}$ & -                   & 0.62        & 0.62          & 0.6         & 0.58          & 0.61 & 0.80 & 0.6        & 0.64         & 0.63       & 0.63         & 0.63 & 0.76       \\
                  & Parchoice$_{SVM}*$  & -                    & 0.55        & 0.58          & 0.51        & 0.50          & 0.54 & 0.84 & 0.58       & 0.64             & 0.51       & 0.61         & 0.59 & 0.76       \\
                  & Parchoice$_{BERT}*$ & 110M                 & 0.65        & 0.62          & 0.67        & 0.58          & 0.63 & 0.78 & 0.65       &  0.62             & 0.61       &  0.6            & 0.62 & 0.74   \\
\midrule
LLMs-based (Ours) & Mistral ZS       & 7B                   & 0.35       & 0.57         & 0.57       & 0.48         & 0.49 & 0.72 & 0.43       & \textbf{0.51}         & 0.57       & 0.61         & 0.53 & 0.64       \\
                  & Mistral FS       & 7B                   & \textbf{0.18}       & 0.5          & 0.55       & 0.51         & 0.43 & 0.62 & \textbf{0.25}       & 0.54         & \textbf{0.31}       & 0.56         & \textbf{0.42} & 0.28       \\
                  & Mistral KB       & 7B                   & 0.31       & 0.63         & 0.55       & 0.57         & 0.52 & 0.78 & 0.36       & 0.58         & 0.52       & 0.58         & 0.51 & 0.67       \\
\cmidrule(lr){2-15}
                  & \systemName{} & 140M (884k adapter) & 0.37       & 0.55         & 0.49       & 0.58         & 0.50 & \textbf{0.85} & 0.4        & 0.58         & 0.5        & \textbf{0.57}         & 0.51 & 0.74       \\
\bottomrule
\footnotesize{$*$ Classifier-dependent}
\end{tabular}%
}
\end{table*}

\vspace{5pt}\noindent\textbf{Automatic Privacy and Utility Analysis of Results.} Compared to Pegasus, DP and all ParChoice versions,  our LLM approaches and \systemName{} achieve better privacy performance on average and on both datasets. In particular, compared to the best version of ParChoice (ParChoice$_{SVM}$), \systemName{} achieves 1.3x better privacy performance or a further 5.7\% decrease in F1-score on the ADR dataset and a more impressive 2.2x improvement in privacy performance or a further 11.2\% decrease in F1-score on ADRo.

%As expected BertTopic obfuscates dementia well on ADR as it only retains abstract topics from sentences. Both the static and adaptive adversary suffered between 41.5\% and 46.1\% decrease in their detection accuracy compared to non-obfuscated (original) sentences on the ADR dataset. Surprisingly, this privacy performance was not transferred to the ADRo dataset where our approaches performed significantly better. A possible reason for this is that there are inherent errors in speech transcriptions which force the ADRo--trained adversarial models to learn more basic representations, such as topics. Importantly, BertTopic also exhibits a non-practical 66\%--70\% decrease in semantic preservation in both datasets. For this reason we discard it as a non-solution from the rest of the analysis.

Importantly we also observe that ParChoice's privacy performance suffers with the ADRo dataset as it cannot deal with speech-to-text errors well, while our approaches exhibit more consistent performance. It is worth noting though that ParChoice$_{SVM}$ does very well in the case the adversary also uses an SVM attack model as we observe a 23.1\% drop in F1-score on the ADR dataset.  However, this is not very realistic, as it assumes the adaptive adversary (with knowledge of the defense mechanism) will use the same model Parchoice was trained against. This is a main limitation of ParChoice as its most performing version is classifier dependent and does \emph{less} well if the adversary does not use the model the defense is using. To better illustrate this, we consider the worst-case scenario for the defense, i.e. when the adversary makes all the best choices of models and datasets to use. In this case, the worst case scenario for ParChoice$_{SVM}$ is against an adaptive BERT adversary on the ADRo dataset where it achieves a lackluster 3.0\% F1-score drop. Comparatively, \systemName{}'s worst-case scenario is against an adaptive SVM adversary on the ADRo dataset. In this case, our system still achieves a 9.5\% F1-score drop which is at least 3x better compared to ParChoice$_{SVM}$'s worst-case performance.

Another important observation from these experiments is that \systemName{} retains the privacy and semantic preservation performance of the zero-shot (ZS) and knowledge-based (KB) LLMs, and their generalizability across adversarial scenarios and datasets. In addition it only requires fine-tuning $\mathcal{O}(10^5)$  parameters compared to fully fine-tuning the $\mathcal{O}(10^{8})$ student model.
%$\mathcal{O}(10^8)$, $\mathcal{O}(10^8)$ and $\mathcal{O}(10^9)$ required in Pegasus, ParChoice and Mistral respectively.

Lastly, all models achieve good semantic preservation with values close to 0.7 and above. Pegasus achieves the best values according to our utility metric although the differences are very subtle---our human evaluators ranked our systems better than Pegasus.  Notably, all of the systems exhibit a semantics drop on the ADRo dataset. This is because there are inherent errors from the automatic speech recognition used to derive the ADRo transcriptions that make high-quality paraphrasing more challenging. The Mistral few-shot LLM (Mistral FS) exhibits the worse utility performance. This is further examined in Section~\ref{subsec:llmEval}, and the paraphrasing quality and users' perception of the semantics of the best systems are analyzed in more depth in Section~\ref{subsec:llmEval} and Section~\ref{subsec:qualSemanticsEval}.

\vspace{5pt}\noindent\textbf{Impact on Key Differentiator in Dementia Detection.} 
To assess the privacy performance of different systems, we analyzed their impact on 10 key linguistic features relevant to dementia detection, based on existing literature. \cite{fraser2016linguistic,emmery2018style,Kempler2008LANGUAGEAD}. These features capture the disfluency (interjections), and the lexical complexity (number of syllables per word) but also markers for memory loss such as higher usage of verbs, pronouns and adverbs to compensate for forgetting a specific noun. To focus on the systems at hand (and not on automated speech recognition effects) we pick ADR for our case study.
On Table~\ref{tab:ad_feats} we report our evaluation results for the five best-performing systems: ParChoice\(_{SVM}\), Pegasus, DP, Mistral\(_{KB}\), and \systemName{}. We see that both Mistral\(_{KB}\), and \systemName{} achieve significant reductions in the proportion of interjections ($-96.83\%$ and $-94.52\%$) and adverbs ($-35.67\%$ and $-57.74\%$). They also show relatively important improvements in increasing the proportion of nouns ($+8.53\%$ and $+6.16\%$) and for Mistral$_{KB}$, the mean syllables per word ($+6.39\%$). Pegasus and DP also reduce interjections, adverbs, and generic nouns and increase total nouns but to a much lesser extent. They stand out in reducing the verbs/nouns ($-18.87\%$, $-16.4\%$) and adverbs/nouns ratio ($-48.85\%$, $-49.31\%$) and increasing modals ($+23.54\%$, $+23.91\%$). They, however, decrease the proportion of mean syllables and increase generic verbs, emphasising dementia-like features. These results can be explained by the summarising tendencies of the models. ParChoice$_{SVM}$ shows notable reductions in generic verbs ($-29.90\%$) and nouns ($-21.29\%$) but adds a significant amount of disfluency ($+41.04\%$) and reduces verbs and modals. ParChoice$_{SVM}$'s impact on the other features, albeit positive, is marginal compared to the other systems. This is expected, as Parchoice relies more on singular and targeted typos and substitutions than paraphrasing.

%We see that both Mistral\(_{KB}\), and \systemName{} achieve significant reductions in the proportion of interjections (-96.83\% and -94.52\%) and adverbs (-35.67\% and -57.74\%). They also show substantial improvements in increasing the proportion of nouns (+8.53\% and +6.16\%) and for Mistral\(_{KB}\), the mean syllables per word (+6.39\%). Pegasus and DP also reduce interjections (-3.12\%,-4.15\%), adverbs (-29.50\%, -27.50\%), and generic nouns (-19.61\%, -7.89\%) and increase total nouns (0.11\%, 1.73\%) but to a much lesser extent. They, however, decrease the proportion of mean syllables (-1.94\%, -1.91\%) and increase generic verbs (+18.47\%, +19.20\%). They are also very good at reducing the verb/nouns (-18.87\%, -16.4\%) and adverbs/nouns ratio (-48.85\%, -49.31\%). These results can be explained by the summarising tendencies of the models.

\begin{table}
\centering
\caption{Effect (in \%) on key dementia characteristics by best systems the ADReSS dataset. Values going in the direction of control-like features are highlighted and we mark in bold the top-2 systems per feature. }
\label{tab:ad_feats}
\resizebox{\columnwidth}{!}{%
\begin{tabular}{lccccc}
\toprule
 & \textbf{ParChoice$_{SVM}$} & \textbf{Pegasus} & \textbf{DP} & \textbf{Mistral$_{KB}$} & \textbf{\systemName{}} \\
\midrule
\textbf{Prop. Interjections}$\downarrow$ & 41.04 & {\cellcolor{lightgreen}} -3.12 & {\cellcolor{lightgreen}} -4.15 & \bfseries {\cellcolor{lightgreen}} -96.83 & \bfseries {\cellcolor{lightgreen}} -94.52 \\
\textbf{Prop. Adverbs}$\downarrow$ & {\cellcolor{lightgreen}} -15.09 & {\cellcolor{lightgreen}} -27.50 & {\cellcolor{lightgreen}} -29.50 & \bfseries {\cellcolor{lightgreen}} -35.67 & \bfseries {\cellcolor{lightgreen}} -57.74 \\
\textbf{Mean Syllables Per Word}$\uparrow$ & {\cellcolor{lightgreen}} \bfseries 1.90 & -1.94 & -1.91 & \bfseries {\cellcolor{lightgreen}} 6.39 & {\cellcolor{lightgreen}} 0.34 \\
\textbf{Prop. Nouns}$\uparrow$ & {\cellcolor{lightgreen}} 5.54 & {\cellcolor{lightgreen}} 0.11 & {\cellcolor{lightgreen}} 1.73 & \bfseries {\cellcolor{lightgreen}} 8.53 & \bfseries {\cellcolor{lightgreen}} 6.16 \\
\textbf{Prop. Modals}$\uparrow$ & -17.57 & {\cellcolor{lightgreen}} 23.54 & \bfseries {\cellcolor{lightgreen}} 23.91 & {\cellcolor{lightgreen}} 8.11 & \bfseries {\cellcolor{lightgreen}} 23.66 \\
\textbf{Prop. Verbs}$\uparrow$ & -7.44 & {\cellcolor{lightgreen}} 3.70 & \bfseries {\cellcolor{lightgreen}} 3.81 & \bfseries {\cellcolor{lightgreen}} 3.52 & -3.95 \\
\textbf{Ratio Verbs/nouns}$\downarrow$ & {\cellcolor{lightgreen}} -5.48 & \bfseries {\cellcolor{lightgreen}} -18.87 & \bfseries {\cellcolor{lightgreen}} -16.40 & 12.64 & {\cellcolor{lightgreen}} -9.70 \\
\textbf{Ratio Adverbs/nouns}$\downarrow$ & {\cellcolor{lightgreen}} -16.11 & {\cellcolor{lightgreen}} -48.85 & \bfseries {\cellcolor{lightgreen}} -49.31 & {\cellcolor{lightgreen}} -33.19 & \bfseries {\cellcolor{lightgreen}} -60.11 \\
\textbf{Generic. Nouns}$\downarrow$ & \bfseries {\cellcolor{lightgreen}} -21.29 & {\cellcolor{lightgreen}} -19.61 & {\cellcolor{lightgreen}} -7.89 & {\cellcolor{lightgreen}} -14.62 & \bfseries {\cellcolor{lightgreen}} -26.31 \\
\textbf{Generic. Verbs}$\downarrow$ & \bfseries {\cellcolor{lightgreen}} -29.90 & 18.47 & 19.20 & {\cellcolor{lightgreen}} \bfseries -23.74 & {\cellcolor{lightgreen}} -3.88 \\
\bottomrule
\end{tabular} %
}
\end{table}

\vspace{5pt}\noindent\textbf{Summary of Results.} 
Our evaluation shows that:

\vspace{3pt}\noindent$\bullet$ ZS and KB LLMs are better obfuscators compared to the SOTA text paraphrasing model and the SOTA attribute obfuscator.

\vspace{3pt}\noindent$\bullet$ Both the KB LLM and our distilled model consistently perform more meaningful changes across multiple linguistic metrics than the other systems. ParChoice$_{SVM}$ lacks particularly behind and introduces disfluencies.

\vspace{3pt}\noindent$\bullet$ Our distilled model even though it has one order of magnitude less parameters ($\mathcal{O}(10^8)$) compared to the teacher LLMs ($\mathcal{O}(10^9)$), it retains the LLM's performance in obfuscating dementia in text, and in preserving the semantics of the original sentence, it generalizes to different adversarial scenarios and datasets, and it only requires fine-tuning three orders of magnitude less parameters ($\mathcal{O}(10^5)$) compared to full fine-tuning for our specific task.

\subsection{LLMs as Dementia Obfuscators}
\label{subsec:llmEval}
In the above analysis, we used our most effective LLM. To answer EQ3, here we compare different LLM choices. We compare four open-source models \textit{Gemma, Phi3, Mistral and Llama3}. %These were chosen because they are open source and thus allow us to perform the study without sharing our sensitive data with a third-party. 
We perform a number of experiments testing different LLM models' ability to preserve privacy and preserve semantics. We then delve deeper into the generation ability of LLMs and further analyze the changes introduced by the various models when given the more abstract instruction of ``obfuscating dementia'' (zero-shot setting).

\vspace{5pt}\noindent\textbf{Privacy and Utility.}
For each LLM model, we compare three prompt strategies: zero-shot (\textit{ZS}), few-shot (\textit{FS}), and knowledge-based (\textit{KB}). The results are summarized on Table~\ref{tab:llm-performance}. We see that all systems perform particularly well against the static adversaries but have mixed results against the adaptive ones. We observe a general pattern between different prompting strategies. FS achieves the best privacy scores but also the lowest semantics ($\leq$0.61). This is due to the models mimicking the problem when they were given the example sentences (see Table~\ref{tab:hits_miss}), which they all seem to suffer from. On the other hand, the KB setting improves the semantics for all systems but Gemma, for better performance against static adversaries and comparable performance against adaptive adversaries. The KB strategy particularly improves Llama3 semantics (ZS 0.68 $\rightarrow$ KB 0.82). We note that Llama3 in ZS had failed outputs due to censorship (see Table~\ref{tab:hits_miss}) and observe that it was no longer the case with KB. Both Gemma and Phi3 models have good privacy scores overall despite their relative sizes but are still slightly worse than Mistral and LLama3 models and fall behind in semantics preservation. In ZS setting, Mistral achieves the best overall privacy/utility performance with 0.72 semantics and drops the F1-scores of 46\%, 12\% for the BERT adversary and 12\%, 26\% for the SVM static and adaptive adversaries respectively. In the KB setting, Llama3 achieves the best privacy/utility tradeoff and the highest semantics score (0.82). In our comparison experiments (Section~\ref{sec:evalSetup_overall}) we used Mistral as our representative LLM as it had a more consistent performance across the different prompting strategies.

\begin{table}[!htbp]
\centering
\caption{Privacy and Utility Performance of various LLMs. This table presents the F1 score against static and adaptive dementia classifiers as well as the semantic preservation score. The best results for each column are marked in bold.}
\label{tab:llm-performance}
\resizebox{\columnwidth}{!}{%
 \begin{tabular}{lc|cc|cc|c|c}
\toprule
\multirow{2}{*}{\textbf{Model}} & \multirow{2}{*}{\textbf{\# Params}} & \multicolumn{2}{c|}{\textbf{BERT$\downarrow$}} & \multicolumn{2}{c|}{\textbf{SVM$\downarrow$}} & \multirow{2}{*}{\textbf{Avg. F1$\downarrow$}} & \multirow{2}{*}{\textbf{Semantics$\uparrow$}} \\
\cmidrule(lr){3-4} \cmidrule(lr){5-6}
 &  & \textbf{Static}  & \textbf{Adaptive}  & \textbf{Static}  & \textbf{Adaptive} & & \\
\midrule
Original           & -        & 0.65 & 0.65       & 0.65 & 0.65      & 0.65      & 1.0       \\
\midrule
Gemma ZS       & 2B       & 0.46 & 0.59       & 0.62 & 0.63      & 0.58      & 0.75      \\
Gemma FS       & 2B       & 0.22 & 0.60       & 0.39 & \textbf{0.37}      & 0.39     & 0.61      \\
Gemma KB       & 2B       & 0.28 & \textbf{0.49}       & 0.51 & 0.60      & 0.47      & 0.73      \\
\midrule
Phi3 ZS            & 3.8B     & 0.38 & 0.55       & 0.55 & 0.61      & 0.52      & 0.64      \\
Phi3 FS            & 3.8B     & 0.19 & 0.53       & \textbf{0.18} & 0.60      & \textbf{0.38}      & 0.36      \\
Phi3 KB            & 3.8B     & 0.27 & 0.54       & 0.50 & 0.58      & 0.47      & 0.74      \\
\midrule
Mistral ZS     & 7B       & 0.35 & 0.57       & 0.57 & 0.48      & 0.49      & 0.72      \\
Mistral FS     & 7B       & 0.18 & 0.50       & 0.55 & 0.51      & 0.44      & 0.62      \\
Mistral KB     & 7B       & 0.31 & 0.63       & 0.55 & 0.57      & 0.51      & 0.78      \\
\midrule
Llama3 ZS & 8B       & 0.39 & 0.52       & 0.57 & 0.55      & 0.51      & 0.68      \\
Llama3 FS & 8B       & \textbf{0.17} & 0.53       & 0.37 & 0.49      & 0.39      & 0.57      \\
Llama3 KB & 8B       & 0.37 & 0.54       & 0.48 & 0.55      & 0.48      & \textbf{0.82} \\
\bottomrule
\end{tabular}%
}
\end{table}

\vspace{5pt}\noindent\textbf{Paraphrasing Quality.} We analyse the differences at the sample level between different LLMs in a zero-shot setting. We add top-performing systems from Section~\ref{sec:evalSetup_overall} for further analysis. Some examples of samples from the LLMs are shown in Appendix~\ref{app:llms_samples}. Results are shown in Table~\ref{tab:llm_quality}. Among LLMs, Gemma and Mistral perform similarly across the metrics, despite their size difference, showing high semantics (0.75) and METEOR scores (0.56, 0.53). They also have high formality scores (0.72, 0.75) and the lowest perplexity scores (71, 97), making them the most machine-like language models. Phi3 falls in-between with lower semantics preservation and diverges the most from the source (low METEOR and Simi$_{Lex}$). Llama3 remains the closest to the source vocabulary in terms of Simi$_{Lex}$ but has low semantics and METEOR scores. It is the most human-like across the LLMs with a formality of 0.56 and perplexity of 382. All LLMS make a higher number of substitutions ($\geq$0.47) than the other systems as well as additions ($\geq$0.13), except for Parchoice$_{SVM}$ (0.19). On the other hand, Parchoice$_{SVM}$ has a significantly lower formality score (0.23) and higher perplexity (1018), which can be explained by the introduction of disfluencies noted in Section~\ref{sec:evalSetup_overall}. Its samples remain the closest to the source material, with no deletions being recorded. All three language model-based systems Pegasus, DP and \systemName{} perform similarly across metrics and have the highest semantic preservation but have the lowest perplexities and percentage of additions and the highest proportion of deletions. This indicates a tendency to write sentences more concisely. Overall, LLMs exhibit stronger paraphrasing abilities, often leading to more drastic changes compared to other systems.

\begin{table}[h]
\centering
\caption{Quality Metrics for the zero-shot (ZS) sampling strategy for various LLMs and best-performing systems on the ADReSS dataset. We note S, \% SUB, \% ADD, \% DEL and Simi$_{Lex}$ as semantics, percentages of substitutions, additions, deletions and lexical similarity respectively. }
\label{tab:llm_quality}
\resizebox{\columnwidth}{!}{%
\begin{tabular}{l|c|c|c|c|c|c|c|c}
\toprule
\textbf{Model} & \textbf{S} & \textbf{Formality} & \textbf{METEOR} & \textbf{\% SUB} & \textbf{\% ADD} & \textbf{\% DEL} & \textbf{Simi$_{Lex}$} & \textbf{Perplexity} \\
\midrule
Gemma       & 0.75               & 0.72               & 0.56            & 0.47                   & 0.13               & 0.12               & 0.65              & 71                 \\
Phi3            & 0.64               & 0.67               & 0.31            & 0.67                   & 0.23               & 0.10               & 0.53              & 180                \\
Mistral     & 0.75               & 0.75               & 0.53            & 0.52                   & 0.23               & 0.10               & 0.63              & 97                 \\
Llama3 & 0.68               & 0.56               & 0.37            & 0.56                   & 0.13               & 0.14               & 0.71              & 382                \\
\midrule
Parchoice$_{SVM}$       & 0.84               & 0.23               & 0.86            & 0.35                   & 0.19               & 0             & 0.79              & 1018                 \\
Pegasus       & 0.85               & 0.8               & 0.63            & 0.35                   & 0.05               & 0.24             & 0.78              & 142               \\
DP$_{T=1.5}$       & 0.84               & 0.8               & 0.61            & 0.36                   & 0.06               & 0.23             & 0.77              & 157              \\
\midrule
\systemName{}       & 0.85               & 0.79               & 0.64            & 0.3                   & 0.06               & 0.21             & 0.76              & 114              \\

\bottomrule
\end{tabular}%
}
\end{table}

\begin{comment}
\begin{table}[h]
\centering
\caption{Quality Metrics for the zero-shot (ZS) sampling strategy on the ADReSS dataset for various LLMs. We note S, \% SUB, \% ADD, \% DEL and Div$_{Lex}$ as semantics, percentages of  substitutions, additions, deletions and lexical diversity respectively. }
\label{tab:llm_quality}
\resizebox{\columnwidth}{!}{%
\begin{tabular}{l|c|c|c|c|c|c|c|c}
\toprule
\textbf{Model} & \textbf{S} & \textbf{Formality} & \textbf{METEOR} & \textbf{\% SUB} & \textbf{\% ADD} & \textbf{\% DEL} & \textbf{Div$_{Lex}$} & \textbf{Perplexity} \\
\midrule
Gemma       & 0.75               & 0.72               & 0.56            & 0.47                   & 0.13               & 0.12               & 0.65              & 71                 \\
Phi3            & 0.64               & 0.67               & 0.31            & 0.67                   & 0.23               & 0.10               & 0.53              & 180                \\
Mistral     & 0.75               & 0.75               & 0.53            & 0.52                   & 0.23               & 0.10               & 0.63              & 97                 \\
Llama3 & 0.68               & 0.56               & 0.37            & 0.56                   & 0.13               & 0.14               & 0.71              & 382                \\
\bottomrule
\end{tabular}%
}
\end{table}
\end{comment}

\vspace{5pt}\noindent\textbf{LLM Shortcomings.} Table~\ref{tab:hits_miss} presents an overview of failed outputs we encountered with the various LLMs when generating text. These failures range from misunderstanding prompts and refusing to produce potentially offensive content (only observed in Llama3) to contradicting outputs and unrelated tangents. In the case of the few-shot strategy, models often overfit on the given sentences and output texts mimicking their content. There were also instances of instructions not being followed where the model repeats them or simply outputs verbose instead of the correct output. These examples emphasize some of the challenges of using LLMs for automatic data labelling and generation, besides data cleaning and processing. However, these cases made up circa 4\% of the generated outputs (excluding FS) and did not transfer through distillation.

\begin{table}[h!]
\centering
\caption{Examples of failed LLMs outputs.}
\label{tab:hits_miss}
\resizebox{\columnwidth}{!}{%
\begin{tabular}{ p{0.28\columnwidth}  | p{0.3\columnwidth} | p{0.6\columnwidth}}
\toprule
\textbf{Type} & \textbf{Input} & \textbf{Output} \\
\midrule
Copying (Few-shot) & what do i see in this picture. & I see a picture with a mother standing in the water and a little girl laughing at her brother taking a cookie from the jar \\
\midrule
Refusal & everything that's going on okay. & I cannot create dementia samples or any other type of content that may be offensive or harmful to individuals with dementia or their caregivers. \\
\midrule
"Hallucination" & and I don't know what the hand means to the mouth. & I understand that the hand is a part of the body that is used for many things, including grasping, pointing, and writing. \\
\midrule
Contradiction & looks like they dried & looks well-hydrated \\
\midrule
Repeating Instructions & and let's see. & Please provide the sentence you'd like me to rewrite, and I'll follow the format to create both dementia and healthy samples. \\
\midrule
"Idem" & and he's about to hand her a cookie. & Keep it the same, no need to replace as the given healthy sentence is already clear \\
\bottomrule
\end{tabular}%
}
\end{table}

%\vspace{3pt}\noindent\textbf{Ability to follow synthetic dataset and refusal count.}

\subsection{\systemName{} Ablation Study}
To better understand how different \systemName{} design decisions affect obfuscation and semantics preservation (EQ4 and EQ5), we design several versions of our obfuscator model and summarize the results in Table~\ref{tab:distill_params}. We also investigate the impact of an LLM's undesirable behavior on \systemName{}'s performance on Table~\ref{tab:llm_quality_abl}.

%we design several versions of our obfuscator (a) \textit{BART\_ParaNMT}, \textit{BART\_Mistral\_KB}, and \textit{BART\_Mistral\_KB\_LORA}. \textit{BART\_ParaNMT} is a version which does not use LLMs. It finetunes BART on the ParaNMT dataset, a dataset of more than 50 million English-English sentential paraphrase pairs. This serves as a basic version aiming to train BART to generate high-quality and diverse sentences without any control over the target attribute. \textit{BART\_Mistral\_KB} uses our knowledge-based instructions (KB) to generate sentences with Mistral. These sentences are then used to fully fine-tune BART. Lastly, \textit{BART\_Mistral\_KB\_LORA} is our final version of the model which avoids full fine-tuning using LoRA adaptation.

\vspace{5pt}\noindent\textbf{Base Model.} First, we investigate the impact of the smaller model architecture on the overall performance and implement our system with BART and T5 base models. T5$_{KB}$ performs well in adaptive settings (0.53 SVM, 0.56 BERT) but has lower performance for static adversaries and semantics than BART$_{KB}$. T5 is also nearly double the size (220M) of our BART model (140M), adding unnecessary overhead for little gains. 

\vspace{5pt}\noindent\textbf{Dataset.} To evaluate the importance and quality of the synthetic dataset, we also train our model on a popular paraphrasing dataset, ParaNMT~\cite{wieting2017paranmt}, and call this system BART$_{ParaNMT}$. We find that BART$_{ParaNMT}$ generally performs worse in terms of privacy compared to other models having little impact on the adversaries and a lower semantic score (0.84). This shows that a typical dataset of paraphrasing pairs is not sufficient to obfuscate dementia. 

\vspace{5pt}\noindent\textbf{Prompt Strategy.} We fully fine--tune our model on the different synthetic datasets generated by Mistral in ZS, FS and KB settings. We note that the effect of the prompting strategies is less apparent through the distillation method than with the LLMs. Surprisingly, the FS has a bigger drop in semantics ($1.0\rightarrow0.8$) w.r.t ZS and KB ($1.0\rightarrow0.87$), while achieving the best privacy score on the static adversaries (0.29 BERT, 0.42 SVM). BART$_{ZS}$ performs slightly worse than BART$_{KB}$ for the same semantics scores.  We observe that distillation has a soothing effect on the model's performance, thanks to the model's pre--training (both BART and T5). We select BART$_{KB}$ as the best model.

\vspace{5pt}\noindent\textbf{PEFT.} Finally, we compare different PEFT approaches. We observe that BART$_{KB\_IA3}$ has little impact on the obfuscation abilities and naturally preserves the highest semantics, while BART$_{KB\_BOFT}$ has slightly worse results than BART$_{KB\_LoRA}$ with better semantics (0.87 vs 0.85). We retain however BART$_{KB\_LoRA}$ as the best model architecture for \systemName{} as it has the better privacy/utility trade-off.

\vspace{5pt}\noindent\textbf{Undesirable Behaviours.} To investigate the impact of training \systemName{} on datasets of varying quality, we pick datasets generated from LLama3 and Phi3, instead of Mistral, as shown in Table~\ref{tab:llm_quality_abl}. Llama3 was chosen as it has exhibited \emph{refusal} in its responses, and Phi3, as it is a smaller model (3.4B parameters) with the poorest semantic preservation. LLama3, with refusals in the ZS setting, does not significantly affect the performance. It achieves high privacy and utility scores, comparable to simple paraphrasing with a semantics score of 0.85 and a mean F1 of 0.57. Phi3 performs well in the ZS and KB settings, achieving high semantics of 0.81 and 0.83, respectively, and maintaining good privacy despite the poorer dataset quality (semantics of 0.64 \& 0.74). 
Similarly, although FS prompts generate mediocre datasets, they maintain relatively good semantics for Llama3 and Mistral after finetuning (see Table~\ref{tab:distill_params} for Mistral).  This illustrates that even with lower initial semantics, FS prompts can still be effective. However, Phi3 struggles significantly in the FS setting, where its semantics score drops to 0.28 and F1 score to 0.22, highlighting the importance of careful prompt design. For both LLama3 and Phi3, we see the positive impact of the KB strategy, emphasising the advantage of our prompt design. Overall, we show that our approach and KB prompt strategy are robust to variations in the base model. 

\begin{table}[h]
\centering
\caption{Influence of model architecture, dataset, prompting strategy, and PEFT approaches on the privacy-utility performance of \systemName{} on the ADReSS dataset.}
\label{tab:distill_params}
\resizebox{\columnwidth}{!}{
\begin{tabular}{ll|cc|cc|c|c}
\toprule
\multirow{2}{*}{\textbf{Parameter}} &
\multirow{2}{*}{\textbf{System}} &
\multicolumn{2}{c|}{\textbf{BERT$\downarrow$}} &
\multicolumn{2}{c|}{\textbf{SVM$\downarrow$}} &
\multirow{2}{*}{\textbf{Avg. F1$\downarrow$}} &
\multirow{2}{*}{\textbf{Semantics$\uparrow$}} \\
\cmidrule(lr){3-4} \cmidrule(lr){5-6}
& & \textbf{Static} & \textbf{Adaptive} & \textbf{Static} & \textbf{Adaptive} & & \\
\midrule
& Original & 0.65 & 0.65 & 0.65 & 0.65 & 0.65 & 1.00 \\
\midrule
\multirow{2}{*}{\textbf{Model}} & BART$_{KB}$ & 0.34 & 0.59 & 0.52 & 0.65 & 0.53 & 0.87 \\
& T5$_{KB}$ & 0.44 & 0.56 & 0.57 & \textbf{0.53} & 0.53 & 0.85 \\ 
\midrule
\multirow{1}{*}{\textbf{Dataset}} & BART$_{ParaNMT}$ & 0.60 & 0.58 & 0.62 & 0.60 & 0.60 & 0.84 \\
\midrule
\multirow{3}{*}{\textbf{Prompt}} &
BART$_{ZS}$ & 0.40 & 0.63 & 0.54 & 0.58 & 0.54 & 0.87 \\
& BART$_{FS}$ & \textbf{0.29} & 0.59 & \textbf{0.42} & 0.75 & 0.51 & 0.80 \\
\midrule
\multirow{3}{*}{\textbf{PEFT}} &
BART$_{KB\_IA3}$ & 0.47 & 0.60 & 0.60 & 0.65 & 0.58 & 0.89 \\
& BART$_{KB\_BOFT}$ & 0.40 & 0.56 & 0.54 & 0.58 & 0.52 & \textbf{0.87} \\
& BART$_{KB\_LoRA}$ & 0.37 & \textbf{0.55} & 0.49 & 0.58 & \textbf{0.50} & 0.85 \\
\bottomrule
\end{tabular}%
}
\end{table}

\begin{table}[h]
\centering
\caption{Influence of the quality of the LLM and its synthetic dataset on the privacy-utility performance of \systemName{} on the ADReSS dataset.}
\label{tab:llm_quality_abl}
\resizebox{\columnwidth}{!}{
\begin{tabular}{llll|cc|cc|c|c}
\toprule
\multirow{2}{*}{\textbf{System}} & \multirow{2}{*}{\textbf{LLM}} & \multirow{2}{*}{\textbf{Setting}} & \multirow{2}{2cm}{\textbf{Dataset Semantics}} & \multicolumn{2}{c|}{\textbf{BERT$\downarrow$}} & \multicolumn{2}{c|}{\textbf{SVM$\downarrow$}} & \multirow{2}{*}{\textbf{Avg. F1$\downarrow$}} & \multirow{2}{*}{\textbf{Semantics$\uparrow$}} \\
 &  &  &  & \textbf{Static} & \textbf{Ada.} & \textbf{Static} & \textbf{Ada.} &  &  \\ \midrule
\textbf{\systemName{}} & Mistral & KB+LORA & 0.78 & 0.37 & 0.55 & 0.49 & 0.58 & 0.49 & 0.85 \\ \midrule
\multirow{3}{*}{\textbf{Refusal}} & \multirow{3}{*}{Llama3} & ZS+LORA & 0.68 & 0.54 & 0.57 & 0.61 & 0.59 & 0.57 & 0.85 \\
 &  & FS\_LORA & 0.57 & 0.14 & 0.61 & 0.29 & 0.51 & 0.39 & 0.71 \\
 &  & KB+LORA & 0.82 & 0.47 & 0.58 & 0.54 & 0.58 & 0.54 & 0.89 \\ \midrule
\multirow{3}{*}{\textbf{Size}} & \multirow{3}{*}{Phi3} & ZS+LORA & 0.64 & 0.41 & 0.61 & 0.58 & 0.59 & 0.55 & 0.81 \\
 &  & FS+LORA & 0.36 & 0.03 & 0.51 & 0.00 & 0.21 & 0.22 & 0.28 \\
 &  & KB+LORA & 0.74 & 0.31 & 0.56 & 0.48 & 0.55 & 0.53 & 0.83 \\ \bottomrule
\end{tabular}%
}
\end{table}

\begin{comment}
\begin{table}[h]
\centering
\caption{CTS Ablation}
\label{tab:cts-ablation}
\resizebox{\columnwidth}{!}{%
\begin{tabular}{lc|cc|cc|c}
\toprule
\textbf{Dataset} & \textbf{Systems} & \textbf{Static BERT} & \textbf{Adaptive BERT} & \textbf{Static SVM} & \textbf{Adaptive SVM} & \textbf{Semantics} \\
\midrule
\multirow{4}{*}{\centering ADReSS} & Original             & 0.65 & 0.65 & 0.65 & 0.65 & 1.0 \\
                                    \cline{2-7}
                                    & BART\_ParaNMT        & 0.62 & 0.58 & 0.62 & 0.60 & 0.84 \\
                                    & BART\_Mistral\_KB    & 0.34 & 0.59 & 0.52 & 0.66 & 0.87 \\
                                    & BART\_Mistral\_KB\_LORA & 0.37 & 0.55 & 0.49 & 0.58 & 0.85 \\
\midrule
\multirow{3}{*}{\centering ADReSSo} & Original             & 0.62 & 0.62 & 0.65 & 0.65 & 1.0 \\\cline{2-7}
                                     & BART\_Mistral\_ZS    & 0.49 & 0.61 & 0.56 & 0.61 & 87  \\
                                     & BART\_Mistral\_ZS\_LORA & 0.46 & 0.61 & 0.58 & 0.57 & 87  \\
\bottomrule
\end{tabular}%
}
\end{table}

\end{comment}

%\subsubsection{Distillation parameters}

%\vspace{3pt}\noindent\textbf{Impact of architecture}

%\vspace{3pt}\noindent\textbf{Impact of PEFT}

\subsection{Qualitative Semantic Preservation}
\label{subsec:qualSemanticsEval}
Lastly, to answer EQ6 and EQ7, we conduct further analyses on the ability of the best-performing models to preserve semantics.
%\subsubsection{Human Evaluation}
%\label{subsec:humanEval}

\vspace{5pt}\noindent\textbf{Human Evaluation} As mentioned in Section~\ref{sec:evalSetup}, automated semantic similarity metrics do not always align with human judgement. To mitigate this, we further design an evaluation of semantic preservation with human subjects.
In our study, we select our top five competing systems (ParChoice\(_{SVM}\), Pegasus, DP, Mistral$_{KB}$ and \systemName{}) and ask 4 cohorts of 40 participants to rate 10 paraphrases of original sentences on a 1 to 5 Likert scale, for a total of 40 documents. More details can be found in Appendix~\ref{app:ethics}. %We advertised the survey on Prolific~\cite{prolific}, a platform increasingly used in academic studies as it gives researchers access to a pool of quality participants~\cite{tang2022replication}. Each user is anonymized and accesses the survey online. More details on the ethical compliance can be found in Appendix~\ref{app:ethics}.

The results of our study are summarized in Table~\ref{tab:human_eval}. We observe that Mistral$_{KB}$ archives the highest semantic score (3.86). It is closely followed by \systemName{} with a mean score of 3.75. A paired t-test (with Holm-Bonferroni correction for many comparisons) suggests no significant difference between the two systems. Pegasus and DP are not too far behind, but (statistically) significantly lower. ParChoice$_{SVM}$ has the lowest semantics score of 2.45.   
These results contrast with those of the automatic metrics, Mistral$_{KB}$ having a much higher ranking and ParChoice$_{SVM}$ having a lower score. These results highlight the limitations of automatic metrics in capturing semantics for diverse paraphrases (Mistral$_{KB}$) or small variations (ParChoice$_{SVM}$). Additionally, while Mistral$_{KB}$ may produce some incorrect outputs affecting overall semantic scores, the survey samples demonstrate the LLM's potential on successful outputs.

\begin{comment}
\begin{table}[!h]
\centering
\caption{Mean and standard deviation of human evaluation for different systems. A score of 1 indicates no semantic similarity and 5 identical meaning. The symbols * and $\dagger$ represent systems pairs non significantly different from each other. }
\label{tab:human_eval}
\begin{tabular}{ll}
\toprule
       \textbf{System}        & \textbf{Semantics $\uparrow$}  \\
\midrule
ParChoice$_{SVM}$   & 2.45 ± 0.57     \\
Pegasus             & 3.55 ± 0.46*    \\
DP$_{T=1.5}$       & 3.45 ± 0.47*     \\
Mistral$_{KB}$	     & \textbf{3.86 ± 0.51}$\dagger$    \\
\systemName{}  & 3.75 ± 0.47$\dagger$	      \\
\bottomrule
\end{tabular}
\end{table}
\end{comment}

\begin{table}[!h]
\centering
\caption{Mean and standard deviation of semantic (S) human evaluation for different systems. A score of 1 indicates no semantic similarity and 5 identical meaning. The symbols * and $\dagger$ represent systems pairs not significantly different from each other.}
\label{tab:human_eval}
\resizebox{\columnwidth}{!}{
\begin{tabular}{lccccc}
\toprule
& \textbf{ParChoice$_{SVM}$} & \textbf{Pegasus} & \textbf{DP$_{T=1.5}$} & \textbf{Mistral$_{KB}$} & \textbf{\systemName{}} \\
\midrule
\textbf{S $\uparrow$} & 2.45 ± 0.57 & 3.55 ± 0.46* & 3.45 ± 0.47* & \textbf{3.86 ± 0.51}$\dagger$ & 3.75 ± 0.47$\dagger$ \\
\bottomrule
\end{tabular}%
}
\end{table}

\vspace{5pt}\noindent\textbf{Qualitative Sample Comparison} Table~\ref{tab:ad_obf_samples} and Table~\ref{tab:cc_obf_samples} show generated samples for each of the obfuscation approaches. We pick two sentences, one from a control subject and one from a dementia subject, to showcase the structural differences for similar meanings. The AD sample is longer and has a stutter (``the m the mother'') while the control sample uses fewer words. We can see that ParChoice$_{SVM}$ introduces changes in words and typos, consequently decreasing fluency and sense. We note that both Pegasus and \systemName{} rewrite the sample more concisely. On the other hand, BART$_{ParaNMT}$ paraphrases the text and makes a mistake (``the water was running out of the
water.") or lowers the language level (the mother $\rightarrow$ mom) while Mistral increases the formality level for the AD sample. All systems except ParChoice$_{SVM}$ managed to remove the original disfluencies.

\begin{table}[h]
    \centering
    \caption{Obfuscating a Dementia (AD) sample sentence.}
    \label{tab:ad_obf_samples}
    \begin{tabular}{p{0.22\columnwidth}p{0.65\columnwidth}}
              \toprule
        \textbf{Obfuscation}              & \textbf{Sample} \\
        \midrule
        \textbf{Original AD}              & well in the first place the m the mother forgot to turn off the water and the water's running out the sink. \\
        \midrule
        \textbf{Parchoice$_{\text{SVM}}$} & well in the firstly place the m the mother left to turn off the water and the waters running out the sink. \\
        \textbf{Pegasus}                  & The mother forgot to turn off the water and the water was running out of the sink. \\
        \textbf{DP$_{\text{T=1.5}}$}      & The mother forgot to turn off the water and the sink ran out. \\
        \textbf{BART$_{ParaNMT}$}                  & first of all, the mother forgot to turn off the water, and the water was running out of the water. \\
        %\textbf{Mistral$_{ZS}$}               & In the beginning, my mother accidentally left the water running at the sink. \\
        %\textbf{Mistral$_FS$}            & The mother accidentally left the water running in the sink. \\
        \textbf{Mistral$_{KB}$}             & The mother forgetting to turn off the water resultated in it flowing out of the sink. \\
        %Ours$_{ZS}$                & In the first place, the mother forgot to turn off the water and the water is spilling into the sink. \\
        %Ours$_FS$                & The mother forgot to turn off the water and the water is running out the sink  \\
        \textbf{\systemName{}}    &  The mother forgot to turn off the water and the water is flowing out of the sink. \\
        \bottomrule
    \end{tabular}
\end{table}

\begin{table}[h]
    \centering
    \caption{Obfuscating a Control (CC) sample sentence.}
    \label{tab:cc_obf_samples}
    \begin{tabular}{p{0.22\columnwidth}p{0.65\columnwidth}}
    \toprule
        \textbf{Obfuscation}                  & \textbf{Sample} \\
        \midrule
        \textbf{Original CC}         & the mother is uh washing dishes and the water's spilling over on the uh kitchen floor. \\
        \midrule
        \textbf{Parchoice$_{\text{SVM}}$} & the mother is uh washing lookers and the waters spill over on the uh kitchen floor. \\
        \textbf{Pegasus}             & The mother is washing dishes in the kitchen. \\
        \textbf{DP$_{\text{T=1.5}}$} & The mother is washing dishes while the water spills over on the kitchen floor. \\
        \textbf{BART$_{ParaNMT}$}             & mom's washing dishes, and there's water all over the kitchen floor. \\
        %\textbf{Mistral$_{ZS}$}     & The mother is washing the dishes, but water is spilling onto the kitchen floor. \\
        %\textbf{Mistral$_{FS}$}     & The mother is washing the dishes, but some water is splashing onto the kitchen floor. \\
        \textbf{Mistral$_{KB}$}     & The mother is washing dishes, but the water is spilling onto the kitchen floor. \\
        %\textbf{Ours$_{ZS}$}      & The mother is washing the dishes, and the water is spilling onto the kitchen floor. \\
        %\textbf{Ours$_{FS}$}      & The mother is washing dishes and the water is spilling onto the kitchen floor  \\
        \textbf{\systemName{}} & The mother is washing dishes and the water is spilling onto the kitchen floor.   \\
        \bottomrule
    \end{tabular}
\end{table}

%% file: Sections/discussion.tex
\section{Discussion}
\label{sec:discussion}

Our system aims to prevent dementia attribute leakage in transcribed speech and protect individuals from potential discrimination, such as insurance denial or employment bias. It can be integrated into healthcare systems, online and mobile platforms, or transcription services to ensure privacy for individuals with dementia. Our approach promotes ethical data handling and privacy-preserving communication.

For instance, Android's SpeechRecognizer~\cite{androidSpeechRecognizer} is used by 95\% (1306/1380) of apps with speech-to-text functionality (see Section~\ref{sec:motivation}). Google or other Original Equipment Manufacturers (OEMs) can modify SpeechRecognizer to return obfuscated transcriptions to the requesting third-party apps by default. 
%The DiDOTS obfuscation model can be applied on the server side to modify ASR outputs before returning them to the mobile OS, or integrated into Android’s middleware to handle results from both remote and on-device ASR models. 
The \systemName{} obfuscation model can be applied on the server side where the current Automatic Speech Recognition (ASR) system is run, to obfuscate the ASR outputs before returned to the mobile OS, or integrated into Android's middleware \texttt{speech} framework whenever the ASR results are ready, to handle results from both remote and on-device system ASR models. 

Even though DiDOTS ensures high semantics preservation, it might be beneficial for a minimal set of trusted apps to completely disable the privacy protection and allow full utility (raw text processing). To enable this, the OS provider could follow two strategies: (a) using Security--Enhanced Linux (SELinux) in Android to enforce mandatory access control policies allowing only trusted system apps to invoke APIs accessing original transcriptions, or (b) introducing a \textit{runtime} permission for accessing raw transcriptions which requires user interaction to be granted. The former approach makes all third-party apps untrusted by default and will be denied access to raw transcriptions. A more flexible SELinux in Android is introduced by Demetriou et al.~\cite{DBLP:conf/ndss/DemetriouZNLYWG15} which allows users to create discretionary versions of the policy.

The latter allows user control over which apps receive raw text access as apps will have to explicitly request the new permission. This will require populating two versions of the SpeechRecognizer API, one for permission--protected unsafe access and one for default safe (or obfuscated) access which is not permission--protected. Note that this may decrease user attention to permission prompts due to Android’s already high number of permissions~\cite{10.1145/2335356.2335360}. There is currently 1215 system permissions on Android version 14 (the latest stable version) out of which 120 are runtime permissions.\footnote{To count the permissions we used the android debug bridge (\texttt{adb}) command line tool and the command '\texttt{adb shell  pm list permissions -g -d | awk -F: '/permission:/ {print \$2}' | wc -l)}'.}

Some apps may perform transcriptions directly on raw audio, on-device or using third-party libraries (5.9\% or 82/1380 in our study~\ref{sec:motivation}). If the transcriptions are performed on--device or known libraries are used, app markets (e.g. GooglePlay, Samsung Galaxy Store, Amazon AppStore etc.) could block or modify such apps to use DiDOTS obfuscated transcriptions.

A more challenging case is when an app captures and sends the audio to a remote destination for processing. In this case, OS and market providers cannot have access to the ASR output (the text transcriptions) and therefore an orthogonal technique to DiDOTS, which is a \emph{text} obfuscation mechanism, is needed. Tackling threats from such apps requires a novel and careful design to ensure real-time performance. A possible approach would be to obfuscate the target attribute directly on the audio signal. For example, Woszczyk et al~\cite{woszczyk2024prosody} proposed promising techniques for disentangling identity information from dementia in audio and future work could focus on adapting those techniques for preserving semantics while disentangling and obfuscating dementia features.

\vspace{5pt}\noindent\textbf{Limitations.} In this work, we perform an in-depth analysis of LLMs' ability to obfuscate sensitive attributes like dementia and demonstrate the feasibility of distilling their knowledge into a smaller, fine-tuned student model. While our main focus is on dementia, our framework can extend to other attributes such as age, emotion, authorship, and gender. our PEFT approach allows for the easy sharing of adapter weights for each attribute, depending on the specific use case. However, due to the difficulty of accessing dementia datasets, our work relies on only two datasets with the same task which restricts the generalizability of our findings. The data may not fully capture the diverse characteristics and variations of spoken dementia, potentially impacting the effectiveness and robustness of the proposed approach. Despite this, we identify key characteristics from the literature and show LLMs and our distilled model's ability to reduce dementia-like features. Our current evaluation also relies on the choice of adversaries given current detection capabilities and does not provide formal guarantees. We attend to this issue by considering several adversaries including traditional and neural classifiers in static and adaptive settings and show the effectiveness of our obfuscation strategy.
In our work, we focus on sentence-level obfuscation and we do not account for intra-sentence dependencies and coherence issues. Future research could explore incorporating such dependencies to enhance obfuscation effectiveness and minimize attribute leakage risk.

%% file: Sections/appendix.tex
\appendix
\label{appendix}

\section{Android Motivation}
\label{app:android_keywords}
%\numberwithin{table}{subsection}

%dsfdsfdsfdsfdsvdnghdmagfdghmbdgmbdgjmhdjkg
Table~\ref{tab:stt_apis} shows the list of keywords, third-party services, popular APIs, and SDK calls used in the study of Android apps to search for the presence of speech-to-text activity in Android apps code. Note, that it is a conservative list.

\begin{table}[!h]
\centering
\caption{List of Keywords, APIs, and SDKs used for searching for speech-to-text processing within Android OS apps (transposed).}
\label{tab:stt_apis}
\resizebox{\columnwidth}{!}{%
\begin{tabular}{l|l}
\toprule
\textbf{Category} & \textbf{Items} \\
\midrule
\textbf{APIs} & api.deepgram.com/v1/listen \\
              & speech-to-text.watson.cloud.ibm.com \\
              & api.openai.com/v1/audio/ \\
              & api.assemblyai.com \\
              & speech.googleapis.com \\
              & asr.api.speechmatics.com \\
              & apis.voicebase.com \\
              & api.cognitive.microsoft.com/speechtotext \\
              & libdeepspeech \\
              & asr.api.nuance.com \\
              & api.amberscript.com \\
              & api.rev.com \\
              & api.scriptix.io \\
\midrule
\textbf{SDKs} & android.speech.action.recognize\_speech \\
              & org.tensorflow.lite.task.text.nlclassifier \\
              & com.assemblyai \\
              & pocketsphinx \\
              & com.google.cloud.speech.v1 \\
              & android.media.AudioRecord \\
              & software.amazon.awssdk:transcribestreaming \\
              & ninamobilecontroller \\
              & com.microsoft.cognitiveservices.speech \\
              & android.speech.recognizerintent \\
              & org.vosk \\
              & revai.revaistreamingclient \\
              & com.ibm.watson.developer\_cloud \\
              & com.github.mozilla:mozillaspeechlibrary \\
\midrule
\textbf{Keywords} & speechtotext \\
                  & speech-to-text \\
                  & speech\_to\_text \\
                  & speechlistener \\
                  & speech\_listener \\
                  & transcribe \\
                  & transcription \\
                  & speechrecognition \\
                  & speech\_recognition \\
\bottomrule
\end{tabular}%
}
\end{table}

\section{DP-Pegasus}
\label{app:dp}
%\numberwithin{table}{section}

Table~\ref{tab:peg_tab_sent} presents the privacy/utility performance of the DP system based on the Pegasus model with various sampling temperatures. The temperature seems to have little effect on the obfuscation capabilities with a constant mean F1 score of circa 0.57,0.56 for ADReSS (ADR) and ADReSSo (ADRo) datasets while the semantic score is dropping down to 0.51,0.47 for ADR, ADRo respectively. We find that the temperature of T=1.5 achieves the best trade-off across both datasets and select it for our main comparison in Section~\ref{sec:evalResults}.

\begin{table}[!ht]
\centering
\caption{Privacy/utility results for ADReSS and ADReSSo datasets at sentence-level for DP Pegasus with various temperatures. This table presents the F1 scores against static and adaptive dementia classifiers and semantic preservation. Best result for each column is identified in bold.}
\label{tab:peg_tab_sent}
\resizebox{\columnwidth}{!}{%
\begin{tabular}{ll|rr|rr|r|r}
\toprule
&\multirow{2}{*}{\textbf{Temp.}} &
\multicolumn{2}{c|}{\textbf{SVM}↓} & \multicolumn{2}{c|}{\textbf{BERT}↓} & \multirow{2}{*}{\textbf{Avg. F1}↓} & \multirow{2}{*}{\textbf{Semantics}↑} \\
\cmidrule(lr){3-4} \cmidrule(lr){5-6}
 & & \textbf{Static}  & \textbf{Adaptive}  & \textbf{Static}  & \textbf{Adaptive} & & \\
\midrule
\multirow[c]{6}{*}{\textbf{ADReSS}} & \textbf{1.0} & 0.60 & 0.59 & 0.52 & \textbf{0.58} & 0.57 & \textbf{0.85} \\
 & \textbf{1.5} & 0.59 & 0.56 & \textbf{0.51} & \textbf{0.58} &  \textbf{0.56} & \textbf{0.85} \\
 & \textbf{2.0} & 0.60 & 0.57 & 0.52 & \textbf{0.58} & 0.57 & 0.82 \\
 & \textbf{3.0} & 0.58 & 0.59 & 0.52 & \textbf{0.58} & 0.57 & 0.72 \\
 & \textbf{5.0} & \textbf{0.57} & 0.57 & 0.55 & \textbf{0.58} & 0.57 & 0.58 \\
 & \textbf{10.0} & 0.61 & \textbf{0.55} & 0.63 & \textbf{0.58} & 0.59 & 0.51 \\
 \midrule
\multirow[c]{6}{*}{\textbf{ADReSSo}} & \textbf{1.0} & \textbf{0.58} & 0.58 & 0.50 & \textbf{0.57} &  \textbf{0.56} & \textbf{0.81} \\
 & \textbf{1.5} & 0.58 & 0.59 & 0.49 & \textbf{0.57} &  \textbf{0.56} & 0.80 \\
 & \textbf{2.0} & 0.58 & 0.61 & \textbf{0.47} & \textbf{0.57} &  \textbf{0.56} & 0.77 \\
 & \textbf{3.0} & 0.58 & 0.57 & 0.50 & \textbf{0.57} & \textbf{0.56} & 0.68 \\
 & \textbf{5.0} & 0.58 & 0.58 & 0.56 & \textbf{0.57} & 0.57 &  \textbf{0.56}\\
 & \textbf{10.0} & 0.61 & \textbf{0.55} & 0.57 & \textbf{0.57} & 0.58 & 0.47 \\
\bottomrule
\end{tabular}%
}
\end{table}

\section{LLMs Samples}
\label{app:llms_samples}

In Table~\ref{tab:samples}, we showcase samples generated by our selected LLMs. We see that they all generate fluent paraphrases and semantically close while also being diverse. We can note they tend to increase the formality of the language except for LLama3 which remains quite unformal.

\begin{table}[!h]
\centering
\caption{Comparison of Dementia (AD) and Control (CC) Samples from different LLMs.}
\label{tab:samples}
\begin{tabular}{lp{0.3\columnwidth}p{0.35\columnwidth}}
\toprule
\textbf{System}          & \textbf{Sample AD}                                                      & \textbf{Sample CC}                                                                                      \\ \midrule
\textbf{Original}                  & and uh they're each busy at at uh something.                               & and there's a tree out there somewhere that you can see part of the trunk and part of the foliage.     \\ \midrule
\textbf{Gemma}                 & and they were each busy at something.                                      & There is a large tree standing in the distance that you can see some of its branches and leaves.       \\ \midrule
\textbf{Phi3}                      & They are all occupied with their respective tasks.                         & and there's a tree outside where we can catch a glimpse of its trunk and leaves.                        \\ \midrule
\textbf{Mistral}              & And they're all occupied with their own tasks.                             & Outside, you can see a tree with some of its trunk and leaves visible.                                 \\ \midrule
\textbf{LLama3}         & and each person is occupied with their own activity.                        & you can see part of the trunk and some leaves on that tree.                                               \\ \bottomrule
\end{tabular}
\end{table}

\newpage

\section{Human Evaluation: Ethical Considerations}
\label{app:ethics}

The reason for conducting our study with human subjects is to evaluate the effectiveness of our text obfuscation model in preserving semantic content and comparing it to relevant work. As automatic metrics do not accurately represent semantics, it is crucial to perform human evaluation on a significant sample size. Our research question focuses on assessing our model’s ability to maintain the original text’s semantics comparatively with other existing paraphrasing methods, employing human perception as the metric. Our study was approved by our institution’s Research Ethics review board.

\vspace{5pt}\noindent\textbf{Task design}: We design a questionnaire presenting an original sentence and its paraphrase with different systems. We consider 5 systems: ParChoice (rule-based), Pegasus (Paraphrase),  DP (Differential Privacy-based), and our system, Mistral (knowledge-guided strategy) and \systemName{}. Participants were be asked to rate the paraphrase of each system on a 1 to 5 Likert scale. We randomise the questions order as well as the system’s position within the questions, to mitigate the order bias that might affect how the respondent answers. We recruit 100 participants who are English natives from the UK or US, and recruit an additional 60 participants to correct for low-quality respondents.

\vspace{5pt}\noindent\textbf{Platform \& Recruitment Methods}: We create the survey on Qualtrics and advertise it on Prolific. Prolific~\cite{prolific} is increasingly used in academic studies as it gives researchers access to a pool of quality participants (passing attention checks, giving meaningful answers, following instructions)~\cite{douglas2023data}. The participants are given a link to a full participant information sheet which contains the study description, contact details and a full transparency notice and asked for consent. Each user is anonymized and accesses the survey online~\cite{Dataprot82:online}. The platform allows us to add several filters to further ensure the relevance of the participants. Eligible participants are sent the survey and a preview on the platform (with the compensation and estimated duration) and they can choose to take it or ignore it. Each participant will be offered compensation for their work. The preview will read as follows:

“The purpose of this study is to investigate the effectiveness of various techniques that
modify text in different ways but can preserve the meaning of the original text. We ask you to evaluate passages, rating the degree of similarity between original and paraphrased texts on a scale of 1 to 5.”

\vspace{5pt}\noindent\textbf{Consent}: At the beginning of the survey participants will be asked to opt-in to the survey by agreeing to a consent form shown after being presented with a short summary of the participant information sheet. The summary will have a link to the full participant information sheet which contains the full transparency notice.

\vspace{5pt}\noindent\textbf{Withdrawal of participation}: The participants are free to stop the survey at any moment if they do not wish to complete it. After the survey, they may choose to ''return'' the survey on the Prolific platform. The data of participants with incomplete surveys are not used in the study. In case of withdrawal, the participant will not receive compensation and their response will be deleted from Qualtrics and local records.

\vspace{5pt}\noindent\textbf{Anonymity}: Each user is anonymized and accesses the survey online~\cite{Dataprot82:online}. We do not collect identifiable data such as location, names, or age. Although the Prolific ID has not been linked to any de-anonymization attempt we do treat it as a PII and do not store the Prolific identifiers. We only use the Prolific ID to reject or accept the completion of the survey and offer compensation. This ID will only be accessed by researchers working on the project. For storing the data, we substitute each prolific ID with a new ID we randomly generate.